\documentclass[twoside,11pt]{article}

\usepackage{blindtext}

\usepackage[preprint]{dmlr2e}

\usepackage{lastpage}

\usepackage{amssymb}
\usepackage{amsfonts} 
\usepackage{booktabs}
\usepackage{comment}
\usepackage{colortbl}
\usepackage{enumitem}
\usepackage{epsfig}
\usepackage{float}
\usepackage[T1]{fontenc}
\usepackage{graphicx}
\usepackage{hyperref} 
\usepackage[utf8]{inputenc} %
\usepackage{microtype}
\usepackage{multicol}
\usepackage{multirow}
\usepackage{nicefrac}
\usepackage{pifont}
\usepackage{tabularx}
\usepackage[normalem]{ulem}
\usepackage{url}
\usepackage{wrapfig}
\usepackage{xcolor}
\usepackage{xspace}

\usepackage{amsmath,amsfonts,bm}

\def\eqref#1{equation~\ref{#1}}

\def\1{\bm{1}}

\DeclareMathAlphabet{\mathsfit}{\encodingdefault}{\sfdefault}{m}{sl}
\SetMathAlphabet{\mathsfit}{bold}{\encodingdefault}{\sfdefault}{bx}{n}

\DeclareMathOperator*{\argmin}{arg\,min}

\newcommand{\orgdataset}{FSRD\xspace}
\newcommand{\ourproblem}{SRSD\xspace}

\def\U{\mathcal{U}}
\def\Ulog{\U_{\log}}
\def\highlight{\bf \cellcolor{lightgray}}
\def\nacell{\multicolumn{1}{c}{N/A}}
\def\nacellmid{\multicolumn{1}{c|}{N/A}}

\newcommand{\xmark}{\color{red}\ding{55}}%
\newcommand{\cmark}{\color{green}\ding{51}}%

\dmlrheading{24}{2024}{1-\pageref{LastPage}}{10/12; Revised 2/19}{2/19}{21-0000}{Matsubara, Chiba, Igarashi and Ushiku} %
\def\openreview{\url{https://openreview.net/forum?id=qrUdrXsiXX}} %

\ShortHeadings{Matsubara, Chiba, Igarashi and Ushiku}{Rethinking Symbolic Regression Datasets and Benchmarks for Scientific Discovery}
\firstpageno{1}

\begin{document}

\title{Rethinking Symbolic Regression Datasets and Benchmarks for Scientific Discovery}

\author{\name Yoshitomo Matsubara\thanks{This work was mainly done while this author was a research intern at OMRON SINIC X Corporation.} \email yoshitom@uci.edu \\
       \addr Amazon Alexa, USA
       \AND
       \name Naoya Chiba\thanks{This work was mainly done while this author was a project researcher at OMRON SINIC X Corporation.}  \email chiba@nchiba.net \\
       \addr Tohoku University, Japan
       \AND
       \name Ryo Igarashi \email ryo.igarashi@sinicx.com \\
       \addr OMRON SINIC X Corporation, Japan
       \AND
       \name Yoshitaka Ushiku \email yoshitaka.ushiku@sinicx.com \\
       \addr OMRON SINIC X Corporation, Japan
}

\editor{Theodoros Rekatsinas}

\maketitle

\begin{abstract}%
This paper revisits datasets and evaluation criteria for Symbolic Regression (SR), specifically focused on its potential for scientific discovery. Focused on a set of formulas used in the existing datasets based on Feynman Lectures on Physics, we recreate 120 datasets to discuss the performance of symbolic regression for scientific discovery (\ourproblem). For each of the 120 \ourproblem datasets, we carefully review the properties of the formula and its variables to design reasonably realistic sampling ranges of values so that our new \ourproblem datasets can be used for evaluating the potential of \ourproblem such as whether or not an SR method can (re)discover physical laws from such datasets. We also create another 120  datasets that contain dummy variables to examine whether SR methods can choose necessary variables only. Besides, we propose to use normalized edit distances (NED) between a predicted equation and the true equation trees for addressing a critical issue that existing SR metrics are either binary or errors between the target values and an SR model's predicted values for a given input. We conduct benchmark experiments on our new \ourproblem datasets using various representative SR methods. The experimental results show that we provide a more realistic performance evaluation, and our user study shows that the NED correlates with human judges significantly more than an existing SR metric.
We publish repositories of our code~\footnote{\url{https://github.com/omron-sinicx/srsd-benchmark} \label{fn:code_repo}} and 240 \ourproblem datasets.\footnote{\url{https://huggingface.co/datasets/yoshitomo-matsubara/srsd-feynman_easy} \label{fn:srsd_easy}}~\footnote{\url{https://huggingface.co/datasets/yoshitomo-matsubara/srsd-feynman_medium} \label{fn:srsd_medium}}~\footnote{\url{https://huggingface.co/datasets/yoshitomo-matsubara/srsd-feynman_hard} \label{fn:srsd_hard}}~\footnote{\url{https://huggingface.co/datasets/yoshitomo-matsubara/srsd-feynman_easy_dummy} \label{fn:srsd_easy_dummy}}~\footnote{\url{https://huggingface.co/datasets/yoshitomo-matsubara/srsd-feynman_medium_dummy} \label{fn:srsd_medium_dummy}}~\footnote{\url{https://huggingface.co/datasets/yoshitomo-matsubara/srsd-feynman_hard_dummy} \label{fn:srsd_hard_dummy}}
\end{abstract}


\section{Introduction}
\label{sec:intro}
Recent advances in machine learning (ML), especially deep learning, have led to the proposal of many methods that can reproduce the given data and make appropriate inferences on new inputs.
Such methods are, however, often black-box, which makes it difficult for humans to understand how they made predictions for given inputs.
This property will be more critical especially when non-ML experts apply ML to problems in their research domains such as physics and chemistry.
Symbolic regression (SR) is the task of producing a mathematical expression (symbolic expression) that fits a given dataset.
SR has been studied in the genetic programming (GP) community~\citep{hoai2002solving,keijzer2003improving,koza2005genetic,johnson2009genetic,uy2011semantically,orzechowski2018we}, and deep learning-based SR methods have been attracting more attention from the machine learning community~\citep{petersen2020deep,landajuela2021discovering,biggio2021neural,valipour2021symbolicgpt,la2021contemporary,kamienny2022end}.
Because of its interpretability, various scientific communities apply SR to advance research in their scientific fields \emph{e.g.}, Physics~\citep{wu2019toward,udrescu2020ai,udrescu2020ai2,kim2020integration,cranmer2020discovering,liu2021aipoincare,liu2021nnphd}, Applied Mechanics~\citep{huang2021ai}, Climatology~\citep{abdellaoui2021symbolic}, Materials~\citep{sun2019data,wang2019symbolic,weng2020simple,loftis2020lattice}, and Chemistry~\citep{batra2020emerging}.
Given that SR has been studied in various communities,~\citet{la2021contemporary} propose a benchmark framework for symbolic regression methods.
In the benchmark study, they combine the Feynman Symbolic Regression Database (\orgdataset)~\citep{udrescu2020ai} and the ODE-Strogatz repository~\citep{strogatz2018nonlinear} and compare a number of SR methods, using a large-scale heterogeneous computing cluster.\footnote{Hosts with 24-28 core Intel(R) Xeon(R) CPU E5-2690 v4 2.60GHz processors and 250GB RAM.}

To discuss the potential of symbolic regression for scientific discovery (\ourproblem), there still remain some issues to be addressed: oversimplified datasets and lack of evaluation metric towards \ourproblem.
For symbolic regression tasks, existing datasets consist of minimum necessary variables and values sampled from limited domains \emph{e.g.}, in range of 1 to 5, and there are no large-scale datasets with reasonably realistic values that capture the properties of the formula and its variables.
Thus, it is difficult to discuss the potential of symbolic regression for scientific discovery using such existing datasets.
For instance, the \orgdataset consists of 120 formulas selected mostly from Feynman Lectures Series\footnote{\citet{udrescu2020ai} extract 20 of the 120 equations as ``bonus'' from other seminal books~\citep{goldstein2002classical,jackson1999classical,weinberg1972gravitation,schwartz2014quantum}.}~\citep{feynman1963feynman1,feynman1963feynman2,feynman1963feynman3} and are core benchmark datasets used in SRBench~\citep{la2021contemporary}.
While the formulas indicate physical laws, variables and constants in each dataset have no physical meanings since the datasets are not designed to discover the physical laws from the observed data in the real world.
(See Section~\ref{subsec:dataset_issues}.)

Moreover, there is a lack of appropriate metrics to evaluate these methods for \ourproblem.
An intuitive approach would be to measure the prediction error or correlation between the predicted values and the target values in the test data, as in standard regression problems.
However, low prediction errors could be achieved even by complex models that differ from the original law.
In addition, SRBench~\citep{la2021contemporary} presents the percentage of agreement between the target and the estimated equations as solution rate.
But in such cases, both 1) equations that do not match at all and 2) that differ by only one term\footnote{If those differ by a constant or scalar, SRBench treats the estimated equation as correct for solution rate.\label{fn:solution_rate}} are equally treated as incorrect.
As a result, it is considered as a coarse-resolution evaluation method for accuracy in \ourproblem, which still needs more discussion towards real-world applications.
A key feature of SR is its interpretability, and some studies~\citep{udrescu2020ai2,la2021contemporary} use complexity of the predicted expression as an evaluation metric (the simpler the better).
However, it is based on a big assumption that a simpler expression may be more likely to be a hidden law in the data (\emph{e.g.}, physical law), which may not be true for \ourproblem.
Therefore, there are no single evaluation metrics proposed to take into account both the interpretability and how close to the true expression the estimated expression is.

To address these issues, we propose new \ourproblem datasets, introduce a new evaluation method, and conduct benchmark experiments using various representative SR baseline methods.
We carefully review and design annotation policies for the new datasets, considering the properties of the physics formulas.
Besides, given that a formula can be represented as a tree structure, we introduce a normalized edit distance on the tree structure to allow quantitative evaluation of predicted formulas that do not perfectly match the true formulas.
Using the proposed \ourproblem datasets and evaluation metric, we perform benchmark experiments with a set of SR baselines and find that there is still significant room for improvements in \ourproblem.
Besides the datasets (Appendix~\ref{app_sec:dataset_additional_info}), we publish our code repository for reproducibility.%

\section{Related Studies}
\label{sec:related_work}
In this section, we briefly introduce related studies focused on 1) symbolic regression for scientific discovery and 2) symbolic regression dataset and evaluation.

\subsection{\ourproblem: Symbolic Regression for Scientific Discovery}
A pioneer study on symbolic regression for scientific discovery is conducted by~\citet{schmidt2009distilling}, who propose a data-driven scientific discovery method.
They collect data from standard experimental systems like those used in undergrad physics education: an air-track oscillator and a double pendulum.
Their proposed algorithm detects different types of laws from the data such as position manifolds, energy laws, and equations of motion and sum of forces laws.

Following the study, data-driven scientific discovery has been attracting attention from research communities and been applied to various domains such as Physics~\citep{wu2019toward,udrescu2020ai,udrescu2020ai2,kim2020integration,cranmer2020discovering,liu2021aipoincare,liu2021nnphd}, Applied Mechanics~\citep{huang2021ai}, Climatology~\citep{abdellaoui2021symbolic}, Materials~\citep{sun2019data,wang2019symbolic,weng2020simple,loftis2020lattice}, and Chemistry~\citep{batra2020emerging}.
These studies leverage symbolic regression in different fields.
While general symbolic regression tasks use synthetic datasets with limited sampling domains for benchmarks, many of the \ourproblem studies collect data from the real world and discuss how we could leverage symbolic regression toward scientific discovery.

While \ourproblem tasks share the same input-output interface with general symbolic regression (SR) tasks (\emph{i.e.}, input: dataset, output: symbolic expression), we differentiate \ourproblem tasks in this study from general SR tasks by whether or not the datasets including true symbolic expressions are created with reasonably realistic assumptions for scientific discovery such as meaning of true symbolic expressions (whether or not they have physical meanings) and sampling domains for input variables.

\subsection{Dataset and Evaluation}
\label{subsec:related_dataset_eval}
For symbolic regression methods, there exist several benchmark datasets and empirical studies.
\orgdataset~\citep{udrescu2020ai} is one of the largest symbolic regression datasets, which consists of 100 physics-inspired equations based on Feynman Lectures on Physics~\citep{feynman1963feynman1,feynman1963feynman2,feynman1963feynman3} and 20 equations from other seminal books~\citep{goldstein2002classical,jackson1999classical,weinberg1972gravitation,schwartz2014quantum}.
By randomly sampling from small ranges of value, they generate the corresponding tabular datasets for the 120 equations.
The ODE-Strogatz repository~\citep{la2016inference} contains two-state dynamic models of first-order, ordinary differential equations sourced from~\citep{strogatz2018nonlinear}.
Inspired by~\citet{hoai2002solving,keijzer2003improving,johnson2009genetic},~\citet{uy2011semantically} suggest 10 different real-valued symbolic regression problems (functions) and create the corresponding dataset (\emph{a.k.a.} Nguyen dataset).
The suggested functions consist of either 1 or 2 variables \emph{e.g.}, $f(x) = x^6 + x^5 + x^4 + x^3 + x^2 + x$ and $f(x, y) = \sin(x) + \sin(y^2)$.
They generate each dataset by randomly sampling 20 - 100 data points.
\citet{la2021contemporary} design a symbolic regression benchmark, named SRBench, and conduct a comprehensive benchmark experiment, using existing symbolic regression datasets such as \orgdataset and ODE-Strogatz repository~\citep{la2016inference}.
In SRBench, SR methods are assessed by 1) an error metric based on squared error between target and estimated values, and 2) solution rate that shows a percentage of the estimated symbolic regression models that match the true models based on \textsf{sympy}~\citep{meurer2017sympy}.
However, these datasets and evaluations are not necessarily designed to discuss the potential of SRSD.
In Sections~\ref{subsec:dataset_issues} and~\ref{subsec:metrics}, we further describe potential issues in prior studies.

\section{Datasets}
\label{sec:datasets}

We summarize issues we found in the existing symbolic regression datasets, and then propose new datasets to address them towards symbolic regression for scientific discovery (\ourproblem).

\subsection{Issues in Existing Datasets}
\label{subsec:dataset_issues}

As introduced in Section~\ref{subsec:related_dataset_eval}, there are many symbolic regression datasets.
However, we consider that novel datasets are required to discuss \ourproblem for the following reasons:

\begin{enumerate}[leftmargin=2em]
    \item {\bf No physical meaning:} Many of the existing symbolic regression datasets~\citep{hoai2002solving,keijzer2003improving,johnson2009genetic,uy2011semantically,trujillo2016neat,jin2019bayesian} are not necessarily physics-inspired, but instead randomly generated \emph{e.g.}, $f(x) = \log(x)$, $f(x, y) = xy + \sin((x - 1) (y - 1))$. To discuss the potential of \ourproblem, we need to further elaborate datasets and evaluation metrics, considering how we would leverage symbolic regression in practice.
    \item {\bf Oversimplified sampling process:} While some of the datasets are physics-inspired such as \orgdataset~\citep{udrescu2020ai} and ODE-Strogatz repository~\citep{la2016inference}, their sampling strategies are very simplified. Specifically, the strategies do not distinguish between constants and variables \emph{e.g.}, speed of light\footnote{We treat speed of light as a constant ($2.998 \times 10^8 ~\mathrm{[m/s]}$).} is treated as a variable and randomly sampled in range of 1 to 5. Besides, most of the sampling domains are far from values we could observe in the real world \emph{e.g}, II.4.23 in Table~\ref{table:easy2} (the vacuum permittivity values are sampled from range of 1 to 5). When sampled ranges of the distributions are narrow, we cannot distinguish Lorentz transformation from Galilean transformation \emph{e.g.} I.15.10 and I.16.6 in Table~\ref{table:medium1}, I.48.2 in Table~\ref{table:medium3}, I.15.3t, I.15.3x, and I.34.14 in Table~\ref{table:hard1}, or the black body radiation can be misestimated to Stephan-Boltzmann law or the Wien displacement law \emph{e.g.} I.41.16 in Table~\ref{table:hard2}.
    \item {\bf Duplicate SR problems:} Due to the two issues above, many of the equations in existing datasets turn out to be duplicate. \emph{e.g.}, as shown in Table~\ref{table:easy1}, $F = \mu N_n$ (I.12.1) and $F = q_2 E$ (I.12.5) in the \emph{original} \orgdataset are considered identical since both the equations are multiplicative and consists of two variables, and their sampling domains (Distributions in Table~\ref{table:easy1}) are exactly the same. For instance, approximately 25\% of the symbolic regression problems in the \emph{original} \orgdataset have 1 - 5 duplicates in that regard.
    \item {\bf Incorrect/Inappropriate formulas:} \orgdataset~\citep{udrescu2020ai} treat every variables as float whereas they should be integer to be physically meaningful. For example, the difference in number of phases in Bragg's law should be integer but sampled as real number (I.30.5 in Table~\ref{table:easy2}). Furthermore, they do not even give special treatment of angle variables (I.18.12, I.18.16, and I.26.2 in Table~\ref{table:easy1}).
    Physically some variables can be negative whereas 
    the \emph{original} \orgdataset only samples positive values (\emph{e.g.} I.8.14 and I.11.19 in Table~\ref{table:medium1}).
    We also avoid using \emph{arcsin}/\emph{arccos} in the equations since the use of \emph{arcsin}/\emph{arccos} in \orgdataset just to obtain angle variables is not experimentally meaningful (I.26.2 in Table~\ref{table:easy1}, I.30.5 in Table~\ref{table:easy2}, and B10 in Table~\ref{table:hard5}). Equations using \emph{arcsin} and \emph{arccos} in the original annotation are I.26.2 (Snell's law), I.30.5 (Bragg's law), and B10 (Relativistic aberration). These are all describing physical phenomena related to two angles, and it is an unnatural deformation to describe only one of them with an inverse function. Additionally, inverse function use implicitly limits the range of angles, but there is no such limitation in the actual physical phenomena.
    \item {\bf Ignoring feature selection:} The existing SR datasets consist of samples using only necessary input variables to symbolically express the true models. \emph{E.g.}, if the true model is $F = \mu N_\text{n}$ (I.12.1) in Table~\ref{table:easy1}, an existing SR dataset would consist of three variables only (\emph{i.e.}, a three-column tabular dataset): two input variables $\mu$ (coefficient of friction), $N_\text{n}$ (normal force), and the target variable $F$ (force of friction). Suppose we do not know the physics law $F = \mu N_\text{n}$. When we observe scenes of the system using some experimental tools, we may measure other input variables 1) ground contact area of the object $a$ and 2) velocity of the object $v$ in addition to $\mu$, $N_\text{n}$, and $F$. When we want to discover the physics law from the observed data points (a five-column tabular dataset), both $a$ and $v$ play the same role as dummy variables and should be excluded through feature selections. SR methods should be able to select the only necessary input variables or features ($\mu$, $N_\text{n}$) from the given data ($\mu$, $N_\text{n}$, $a$, $v$, and $w$), but we cannot discuss such robustness of SR methods, using the existing SR datasets.
\end{enumerate}

\begin{table}[t]
    \caption{SR dataset comparisons with respect to issues summarized in Section~\ref{subsec:dataset_issues}. ({\xmark}: not addressed, {\cmark}: addressed)}
    \label{table:dataset_comparisons}
    \vspace{-1em}
    \def\arraystretch{1.1}
    \begin{center}
        \bgroup
        \small
        \setlength{\tabcolsep}{0.5em}
        \begin{tabular}{l|rccccc}
            \toprule
            \multicolumn{1}{c|}{\bf Dataset} & \multicolumn{1}{c}{\bf \#problems} & \multicolumn{1}{c}{\bf Issue 1} & \multicolumn{1}{c}{\bf Issue 2} & \multicolumn{1}{c}{\bf Issue 3} & \multicolumn{1}{c}{\bf Issue 4} & \multicolumn{1}{c}{\bf Issue 5} \\
            \midrule
            \citep{hoai2002solving} & 4 & \xmark & \xmark & \cmark & \cmark & \xmark \\
            \citep{keijzer2003improving} & 15 & \xmark & \xmark & \xmark & \cmark & \xmark \\
            \citep{johnson2009genetic} & 7 & \xmark & \xmark & \cmark & \cmark & \xmark \\
            \citep{uy2011semantically} & 10 & \xmark & \xmark & \cmark & \cmark & \xmark \\
            \citep{trujillo2016neat} & 9 & \xmark & \xmark & \cmark & \cmark & \xmark \\
            \citep{la2016inference} & 10 & \cmark & \xmark & \cmark & \cmark & \xmark \\
            \citep{jin2019bayesian} & 6 & \xmark & \xmark & \cmark & \cmark & \xmark \\
            \citep{udrescu2020ai} & 120 & \cmark & \xmark & \xmark & \xmark & \xmark \\
            \midrule
            {\bf Ours: \ourproblem-Feynman} & 240 & \cmark & \cmark & \cmark & \cmark & \cmark \\
            \bottomrule
        \end{tabular}
        \egroup
    \end{center}
\end{table}

\subsection{Proposed \ourproblem Datasets}
\label{subsec:our_datasets}

We address the issues in existing datasets above by proposing new \ourproblem datasets based on the equations used in the \orgdataset~\citep{udrescu2020ai}. 
Section~\ref{subsec:dataset_issues} and Table~\ref{table:dataset_comparisons} summarize the differences between the \orgdataset and our \ourproblem datasets.
Our annotation policy is carefully designed to simulate typical physics experiments so that the \ourproblem datasets can engage studies on symbolic regression for scientific discovery in the research community.

\subsubsection{Annotation policy}
\label{subsubsec:annotation_policy}
We thoroughly revised the sampling range for each variable from the annotations in the \orgdataset~\citep{udrescu2020ai}.
First, we reviewed the properties of each variable and treated physical constants (\emph{e.g.}, speed of light, gravitational constant) as constants while such constants are treated as variables in the original \orgdataset datasets ({\bf Issues 1, 4}).
As shown in Table~\ref{table:easy1}, it also makes I.12.1 and I.12.5 two separate problems in \ourproblem datasets while these two problems in the original \orgdataset are duplicates because both the problems share the identical symbolic expressions and sampling ranges.
Next, we defined sampling ranges in SI units to correspond to each typical physics experiment to confirm the physical phenomenon for each equation ({\bf Issues 2, 3, 4}).
We referenced~\citep{feynman1963feynman1,feynman1963feynman2,feynman1963feynman3,nationalhandbook2022} to understand the context in which each formula appeared in our datasets.
Taking mass as an example, it can be the mass of the Earth, an atom, or something else, depending on the context of each formula.
In cases where a specific experiment is difficult to be assumed, ranges were set within which the corresponding physical phenomenon can be seen.
Generally, the ranges are set to be sampled on log scales within their orders as $10^2$ in order to take both large and small changes in value as the order changes.
Variables such as angles, for which a linear distribution is expected are set to be sampled uniformly.
In addition, variables that take a specific sign were set to be sampled within that range.
Tables~\ref{table:easy1} and~\ref{table:easy2} --~\ref{table:hard5} show the detailed comparisons between the original \orgdataset and our proposed \ourproblem datasets.
We also build another 120 \ourproblem datasets, which contain dummy variables to discuss the robustness of SR methods against dummy variables ({\bf Issue 5}), and there will be 240 proposed datasets in total.\footnote{We provide scripts to generate the datasets as part of our code repository.}
See Section~\ref{subsec:dummy_variables} for the detail of the datasets with dummy variables.

\begin{table*}[!h]
    \caption{\uline{Easy set} of our proposed datasets (part 1). C: Constant, V: Variable, F: Float, I: Integer, P: Positive, N: Negative, NN: Non-Negative, $\U$: Uniform distribution, $\Ulog$: Log-Uniform distribution. Other 110 datasets are summarized in Tables~\ref{table:easy2} -~\ref{table:hard5}.}
    \label{table:easy1}
    \def\arraystretch{1.2}
    \begin{center}
        \bgroup
        \setlength{\tabcolsep}{0.3em}
        \scriptsize
        \begin{tabular}{lcllcccc} 
            \toprule
            \multicolumn{1}{c}{\multirow{2}{*}{\bf Eq. ID}} & \multicolumn{1}{c}{\multirow{2}{*}{\bf Formula}} & \multicolumn{2}{c}{\multirow{2}{*}{\bf Symbols}} & \multicolumn{2}{c}{\bf Properties} & \multicolumn{2}{c}{\bf Distributions}
            \\ %
            & & & & Original & Ours & Original & Ours \\\midrule
            \multirow{3}{*}{I.12.1} & \multirow{3}{*}{$F = \mu N_\text{n}$} & $F$ & Force of friction & V, F, P & V, F, P & N/A & N/A\\
             &  & $\mu$ & Coefficient of friction & V, F, P & V, F, P & $\U(1, 5)$ & $\Ulog(10^{-2}, 10^{0})$\\
             &  & $N_\text{n}$ & Normal force & V, F, P & V, F, P & $\U(1, 5)$ & $\Ulog(10^{-2}, 10^{0})$\\\hline
            \multirow{4}{*}{I.12.4} & \multirow{4}{*}{$E = \frac{q_1}{4 \pi \epsilon r^2}$} & $E$ & Magnitude of electric field & V, F, P & V, F & N/A & N/A\\
             &  & $q_1$ & Electric charge & V, F, P & V, F & $\U(1, 5)$ & $\Ulog(10^{-1}, 10^{1})$\\
             &  & $r$ & Distance & V, F, P & V, F, P & $\U(1, 5)$ & $\Ulog(10^{-1}, 10^{1})$\\
             &  & $\epsilon$ & Vacuum permittivity & V, F, P & C, F, P & $\U(1, 5)$ & $8.854 \times 10^{-12}$\\\hline
            \multirow{3}{*}{I.12.5} & \multirow{3}{*}{$F = q_2 E$} & $F$ & Force & V, F, P & V, F & N/A & N/A\\
             &  & $q_2$ & Electric charge & V, F, P & V, F & $\U(1, 5)$ & $\Ulog(10^{-1}, 10^{1})$\\
             &  & $E$ & Electric field & V, F, P & V, F & $\U(1, 5)$ & $\Ulog(10^{-1}, 10^{1})$\\\hline
            \multirow{4}{*}{I.14.3} & \multirow{4}{*}{$U = m g z$} & $U$ & Potential energy & V, F, P & V, F & N/A & N/A\\
             &  & $m$ & Mass & V, F, P & V, F, P & $\U(1, 5)$ & $\Ulog(10^{-2}, 10^{0})$\\
             &  & $g$ & Gravitational acceleration & V, F, P & C, F, P & $\U(1, 5)$ & $9.807 \times 10^{0}$\\
             &  & $z$ & Height & V, F, P & V, F & $\U(1, 5)$ & $\Ulog(10^{-2}, 10^{0})$\\\hline
            \multirow{3}{*}{I.14.4} & \multirow{3}{*}{$U = \frac{k_\text{spring} x^2}{2}$} & $U$ & Elastic energy & V, F, P & V, F, P & N/A & N/A\\
             &  & $k_\text{spring}$ & Spring constant & V, F, P & V, F, P & $\U(1, 5)$ & $\Ulog(10^{2}, 10^{4})$\\
             &  & $x$ & Position & V, F, P & V, F & $\U(1, 5)$ & $\Ulog(10^{-2}, 10^{0})$\\\hline
            \multirow{4}{*}{I.18.12} & \multirow{4}{*}{$\tau = r F \sin\theta$} & $\tau$ & Torque & V, F & V, F & N/A & N/A\\
             &  & $r$ & Distance & V, F, P & V, F, P & $\U(1, 5)$ & $\Ulog(10^{-1}, 10^{1})$\\
             &  & $F$ & Force & V, F, P & V, F, P & $\U(1, 5)$ & $\Ulog(10^{-1}, 10^{1})$\\
             &  & $\theta$ & Angle & V, F, NN & V, F, NN & $\U(0, 5)$ & $\U(0, 2 \pi)$\\\hline
            \multirow{5}{*}{I.18.16} & \multirow{5}{*}{$L = m r v \sin\theta$} & $L$ & Angular momentum & V, F & V, F & N/A & N/A\\
             &  & $m$ & Mass & V, F, P & V, F, P & $\U(1, 5)$ & $\Ulog(10^{-1}, 10^{1})$\\
             &  & $r$ & Distance & V, F, P & V, F, P & $\U(1, 5)$ & $\Ulog(10^{-1}, 10^{1})$\\
             &  & $v$ & Velocity & V, F, P & V, F, P & $\U(1, 5)$ & $\Ulog(10^{-1}, 10^{1})$\\
             &  & $\theta$ & Angle & V, F, P & V, F, NN & $\U(1, 5)$ & $\U(0, 2 \pi)$\\\hline
            \multirow{3}{*}{I.25.13} & \multirow{3}{*}{$V = \frac{q}{C}$} & $V$ & Voltage & V, F, P & V, F & N/A & N/A\\
             &  & $q$ & Electric charge & V, F, P & V, F & $\U(1, 5)$ & $\Ulog(10^{-5}, 10^{-3})$\\
             &  & $C$ & Electrostatic Capacitance & V, F, P & V, F, P & $\U(1, 5)$ & $\Ulog(10^{-5}, 10^{-3})$\\\hline
            \multirow{3}{*}{I.26.2} & \multirow{3}{*}{$n = \frac{\sin\theta_1}{\sin\theta_2}$} & $n$ & Relative refractive index & V, F, NN & V, F, P & $\U(0, 1)$ & N/A\\
             &  & $\theta_1$ & Refraction angle 1 & V, F & V, F, NN & N/A & $\U(0, \frac{\pi}{2})$\\
             &  & $\theta_2$ & Refraction angle 2 & V, F, P & V, F, NN & $\U(1, 5)$ & $\U(0, \frac{\pi}{2})$\\\hline
            \multirow{4}{*}{I.27.6} & \multirow{4}{*}{$f = \frac{1}{\frac{1}{d_1}+\frac{n}{d_2}}$} & $f$ & Focal length & V, F, P & V, F, P & N/A & N/A\\
             &  & $d_1$ & Distance & V, F, P & V, F, P & $\U(1, 5)$ & $\Ulog(10^{-3}, 10^{-1})$\\
             &  & $n$ & Refractive index & V, F, P & V, F, P, & $\U(1, 5)$ & $\Ulog(10^{-1}, 10^{1})$\\
             &  & $d_2$ & Distance & V, F, P & V, F, P & $\U(1, 5)$ & $\Ulog(10^{-3}, 10^{-1})$\\
            \bottomrule
        \end{tabular}
        \egroup
    \end{center}
\end{table*}

\subsubsection{Complexity-aware Dataset Categories}
\label{subsubsec:dataset_categories}
While the proposed datasets consist of 120 different problems, there will be non-trivial training cost required to train a symbolic regression model for all the problems individually~\citep{la2021contemporary} \emph{i.e.}, there will be 120 separate training sessions to assess the symbolic regression approach.
To allow more flexibility in assessing symbolic regression models for scientific discovery, we define three clusters of the proposed datasets based on their complexity: \emph{Easy}, \emph{Medium}, and \emph{Hard} sets, which consist of 30, 40, and 50 different problems respectively.

\begin{figure}[t]
    \centering
    \includegraphics[width=0.8\linewidth]{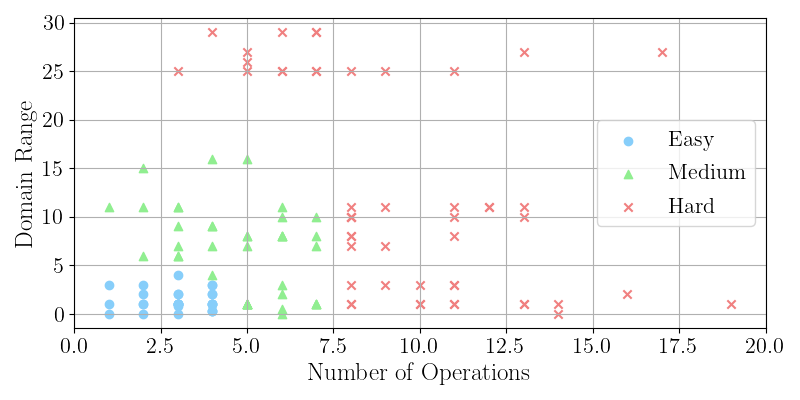}
    \vspace{-1.5em}
    \caption{Distribution map of three subsets for our \ourproblem datasets with respect to our complexity metrics of SR problem. Data points at top right/bottom left indicate more/less complex problems.}
    \label{fig:dataset_scatter}
\end{figure}

We define the complexity of a problem, using the number of operations to represent the true equation tree and range of the sampling domains.
The former measures how many mathematical operations compose the true equation such as \emph{add}, \emph{mul}, \emph{pow}, \emph{exp}, and \emph{log} operations (see Fig.~\ref{fig:eq_preprocess}).
The latter considers magnitude of sampling distributions (\emph{Distributions} column in Tables~\ref{table:easy1} and~\ref{table:easy2} --~\ref{table:hard5}) and increases the complexity when sampling values from wide range of distributions.
We define the domain range as follows:
\begin{equation}
    f_\text{range}\left(\mathcal{S}\right) = \left| \log_{10} \left| \max_{s \in \mathcal{S}} s - \min_{s \in \mathcal{S}} s \right| \right|,
    \label{eq:domain_range}
\end{equation}
\noindent where $\mathcal{S}$ indicates a set of sampling domains (\emph{distributions}) for a given symbolic regression problem.
Using these two metrics, we define \emph{Easy}, \emph{Medium}, or \emph{Hard} sets as illustrated in Fig.~\ref{fig:dataset_scatter}.

These clusters represent problem difficulties at high level.
For instance, these subsets will help the research community to shortly tune and/or perform sanity-check new approaches on the \emph{Easy} set (30 problems) instead of using the whole datasets (120 problems).
Figure~\ref{fig:dataset_scatter} shows the three different distribution maps of our proposed datasets.

\subsection{Introducing Dummy Variables to Our \ourproblem Datasets}
\label{subsec:dummy_variables}

As pointed out in Section~\ref{subsec:dataset_issues}, existing SR datasets such as the \orgdataset consist of only necessary variables to express the predicted equation.
However, there may be irrelevant features (input variables) in the observed samples for real-world applications, and then SR methods should detect and exclude such input variables from their predicted solutions (equations).
The existing SR datasets are not suitable for benchmarking SR methods from the aspect, especially for \ourproblem problems, and thus we introduce dummy variables to our \ourproblem datasets:
\begin{enumerate}[leftmargin=2em]
    \item Given an \ourproblem dataset (input variables + target variable), randomly choose $k_\text{dummy}$, the number of dummy variables to be introduced, from $\{1, 2, 3\}$.
    \item For each of the $k_\text{dummy}$ dummy variables,
    \begin{enumerate}[label*=\arabic*.,leftmargin=2em]
        \item randomly choose the index of the dummy variable (column index of the dummy variable in the resulting tabular dataset),
        \item randomly determine (with a probability of 50\%) whether or not the dummy variable can be sampled from negative sampling range, and
        \item randomly choose $s$ from $\{-32, -31, -30, \dots, 30, 31, 32\}$ and sample $N$ values from $\Ulog(10^{s - 1}, 10^{s + 1})$ for the dummy variable, where $N$ is the number of samples in the given \ourproblem dataset.
    \end{enumerate}
\end{enumerate}

We apply the above procedure to each of the 120 \ourproblem-Feynman datasets independently.
Thus, each of the 120 new \ourproblem datasets (``\ourproblem-Feynman + Dummy Variables'' in Table~\ref{table:baseline_results_w_dummy_vars}) will have a different configuration: number of dummy variables, indices of the dummy variables, and their sampling ranges.
Table~\ref{table:nums_dummy_vars} summarizes how many random dummy variables were introduced to which datasets (equations).

\begin{table}[t]
    \caption{Equation IDs and numbers of dummy variables introduced to our \ourproblem datasets.}
    \label{table:nums_dummy_vars}
    \def\arraystretch{1.1}
    \begin{center}
        \bgroup
        \small
        \begin{tabularx}{\textwidth}{l|XXX} 
            \toprule
            \multicolumn{1}{c|}{\bf Group} & \multicolumn{1}{c}{\bf 1 dummy variable}  & \multicolumn{1}{c}{\bf 2 dummy variables}  & \multicolumn{1}{c}{\bf 3 dummy variables} \\
            \midrule
            Easy & I.12.1, I.12.4, I.12.5, I.18.12, I.25.13, I.47.23 & I.14.3, I.18.16, I.43.16, II.3.24, II.8.31, II.10.9, II.13.17, II.15.5, II.27.18, III.7.38, III.12.43 & I.14.4, I.26.2, I.27.6, I.30.5, II.2.42, II.4.23, II.15.4, II.27.16, II.34.11, II.34.29b, II.38.3, II.38.14, III.15.27 \\
            \cline{1-4}
            Medium & I.10.7, I.12.2, I.13.12, I.16.6, I.32.5, I.43.31, II.11.3, II.34.2, II.34.29a, III.14.14, III.15.14, B8 & I.11.19, I.12.11, I.13.4, I.15.10, I.18.4, I.24.6, I.34.8, I.38.12, I.39.11, I.43.43, I.48.2, II.6.11, II.21.32, II.34.2a, III.4.32, III.13.18, III.15.12, III.17.37 & I.8.14, I.29.4, I.34.10, I.34.27, I.39.10, II.8.7, II.37.1, III.8.54, III.19.51, B18 \\
            \cline{1-4}
            Hard & I.15.3x, I.30.3, II.6.15a, II.11.17, II.11.28, II.13.23, II.13.34, II.24.17, B1, B6, B12, B16, B17 & I.6.20, I.6.20b, I.9.18, I.15.3t, I.29.16, I.34.14, I.39.22, I.44.4, II.11.20, II.11.27, II.35.18, III.9.52, III.10.19, III.21.20, B2, B3, B7, B9 & I.6.20a, I.32.17, I.37.4, I.40.1, I.41.16, I.50.26, II.6.15b, II.35.21, II.36.38, III.4.33, B4, B5, B10, B11, B13, B14, B15, B19, B20 \\
            \bottomrule
        \end{tabularx}
        \egroup
    \end{center}
\end{table}

\section{Benchmark}
\label{sec:benchmark}

Besides the conventional metrics, we propose a new metric to discuss the performance of symbolic regression for scientific discovery in Section~\ref{subsec:metrics}.
Following the set of metrics, we design an evaluation framework of symbolic regression for scientific discovery, hoping that the proposed \ourproblem benchmark helps non-ML experts choose SR methods for their problems.

\subsection{Metrics}
\label{subsec:metrics}

In general, it would be difficult to define ``accuracy'' of symbolic regression models since we will compare its estimated equation to the ground truth equation and need criteria to determine whether or not it is ``correct''.
\citet{la2021contemporary} suggest a reasonable definition of symbolic solution, which is designed to capture symbolic regression models that differ from the true model by a constant or scalar.\footnote{Code in~\citet{la2021contemporary} ignores coefficient terms whose absolute values are less than $10^{-4}$.}
They also use $R^2$ score (Eq.~\ref{eq:r2_score}) and define as accuracy the percentage of symbolic regression problems that a model meets $R^2 > \tau$, where $\tau$ is a threshold \emph{e.g.}, $\tau = 0.999$ in~\citep{la2021contemporary}:
\begin{equation}
    R^2 = 1 - \frac{\sum_j^{N} \left( f_\text{pred}\left( X_{\text{test}, j} \right) - f_\text{true}\left( X_{\text{test}, j} \right) \right)^2}{\sum_k^{N} \left( f_\text{true}\left( X_{\text{test}, k} \right) - \bar{y} \right)^2},
    \label{eq:r2_score}
\end{equation}
\noindent where $N$ indicates the number of test samples (\emph{i.e.}, the number of rows in the test dataset $X_\text{test}$), and $X_{\text{test}, i}$ indicates the $i$-th test sample.
$\bar{y}$ is a mean of target outputs produced by $f_\text{true}$.
$f_\text{pred}$ and $f_\text{true}$ are a trained SR model and a true model, respectively.
However, these two metrics are still binary (correct or not) or require a threshold and do not explain how \emph{structurally close} to the true equation the estimated one is.
While a key feature of symbolic regression is its interpretability, there are no single evaluation metrics to take into account both the interpretability and how close to the true expression the estimated expression is.

\begin{figure*}[t]
    \centering
    \includegraphics[width=0.8\linewidth]{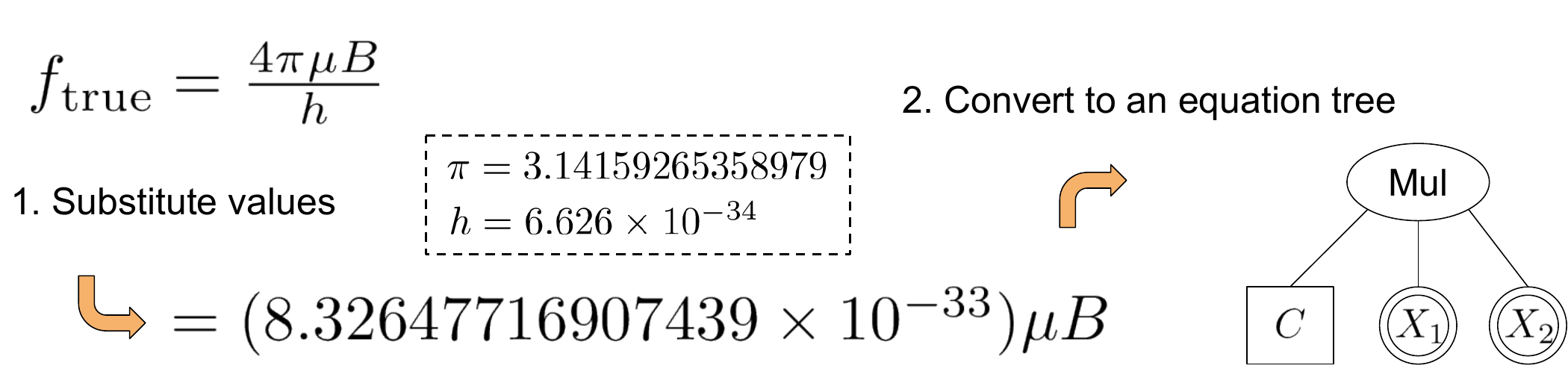}
    \vspace{-1em}
    \caption{Example of preprocessing a true equation (III.7.38 in Table~\ref{table:easy2}) in evaluation session. When converting to an equation tree, we replace constant values and variables with specific symbols \emph{e.g.}, $8.32647716907439 \times 10^{-33} \rightarrow C, \mu \rightarrow X_1, B \rightarrow X_2$.}
    \label{fig:eq_preprocess}
\end{figure*}

To offer more flexibility and assess estimated equations in such a way, we propose use of edit distance between estimated and ground truth equations, processing equations as trees.
Although edit distance has been employed in different domains such as machine translation~\citep{przybocki2006edit} (text-based edit distance), its primary use has been to study the search process for genetic programming approaches~\citep{o1997using,burke2002advanced,nakai2013acquisition}.
Different from prior work, we propose a use of tree-based edit distance as a new metric of solution quality for \ourproblem.
For a pair of two trees, edit distance computes the minimum cost to transform one to another with a sequence of operations, each of which either 1) \emph{inserts}, 2) \emph{deletes}, or 3) \emph{renames} a node.
In this study, a node can be either a mathematical operation (\emph{e.g.}, \emph{add}, \emph{exp} as symbols), a variable symbol, or a constant symbol.
For the detail of the algorithm, we refer readers to~\citep{zhang1989simple}.

As illustrated in Fig.~\ref{fig:eq_preprocess}, we preprocess equations by 1) substituting constant values \emph{e.g.}, $\pi$ and Planck constant to the expression, and 2) converting the resulting expression to an equation tree that represents the preorder traversal of the equation with simplified symbols.
It should be worth noting that before generating the equation tree, we simplify and convert equations to floating-point approximations\footnote{\url{https://github.com/omron-sinicx/srsd-benchmark/blob/main/eq_comparator.py}} by \textsf{sympy}~\cite{meurer2017sympy}, a Python library for symbolic mathematics.
It helps us consistently map a given equation to the unique equation tree and compute edit distance between the true and estimated equation trees since our evaluation interest is in simplified expressions of the estimated equations rather than how SR models produced the equations.
For instance, ``$x + x + x$'', ``$4 * x - x$'', and ``$x + 2 * x$'' will be simplified by \textsf{sympy} to ``$3 * x$'' and considered identical.

For edit distance, we use a method proposed by~\citet{zhang1989simple}.
Given that the range of edit distance values depends on complexity of equations, we normalize the distance in range of 0 to 1 as
\begin{equation}
    \overline{d}(f_\text{pred}, f_\text{true}) = \min\left(1, \frac{d\left(f_\text{pred}, f_\text{true}\right)}{\left|f_\text{true}\right|}\right),
    \label{eq:norm_edit_dist}
\end{equation}
\noindent where $f_\text{pred}$ and $f_\text{true}$ are estimated and true equation trees, respectively.
$d(f_\text{pred}, f_\text{true})$ is an edit distance between $f_\text{pred}$ and $f_\text{true}$.
$|f_\text{true}|$ indicates the number of the tree nodes that compose an equation $f_\text{true}$.
We note that this metric is designed to capture similarity between estimated and true equations, thus coefficient values themselves (\emph{e.g.}, value of $C$ in Fig.~\ref{fig:eq_preprocess}) should not be important.

\subsection{Evaluation Framework}
\label{subsec:eval_framework}

For each problem, we use the validation tabular dataset and choose the best trained SR model $f_\text{pred}^{*}$ from $\mathcal{F}$, a set of the trained models by a given method respect to Eq. (\ref{eq:relative_error}) 
\begin{equation}
    f_\text{pred}^{*} = \argmin_{f_\text{pred} \in \mathcal{F}} \frac{1}{n}\sum_{i=1}^n\left|\frac{f_\text{pred}(X_{\text{val}, i}) - f_\text{true}(X_{\text{val}, i})}{f_\text{true}(X_{\text{val}, i})}\right|^2,
    \label{eq:relative_error}
\end{equation}
\noindent where $X_{\text{val}, i}$ indicates the $i$-th row of the validation tabular dataset $X_\text{val}$.

In our \ourproblem-Feynman, we provide true equations as part of test datasets besides tabular data for benchmark purposes.
In real-world applications, however, only observed samples (\emph{e.g.}, tabular data) are available for training, validation, and test.
Notice that similar to the solution rate proposed in~\citep{la2021contemporary}, normalized edit distance (NED) requires both the predicted and true equations.
For this reason, we use the geometrical distance between predicted values against a validation tabular dataset to choose the best model obtained through hyperparameter tuning.
Using the best model per method, we compute the normalized edit distance to assess the given method.

\section{Experiments}
\label{sec:experiments}

\subsection{Baseline Methods}
\label{subsec:baselines}
For baselines, we use the five best symbolic regression methods in SRBench~\citep{la2021contemporary}.
Specifically, we choose gplearn~\citep{koza2005genetic}, AFP~\citep{schmidt2011age}, AFP-FE~\citep{schmidt2009distilling}, AI Feynman (AIF)~\citep{udrescu2020ai2}, and DSR~\citep{petersen2020deep}, referring to the rankings of solution rate for the \orgdataset datasets in their study.
We note that~\citet{la2021contemporary} also benchmark symbolic regression methods for black-box problems,\footnote{Since the set of the black-box problems is not either physics-inspired or aligned with our scope of scientific discovery (\emph{e.g.}, car price estimation from car width, height, length, etc), we do not use the datasets in this study.} whose true symbolic expressions are unknown, and other symbolic regression methods \emph{e.g.}, Operon~\citep{kommenda2020parameter}, SBP-GP~\citep{virgolin2019linear}, FEAT~\citep{la2018learning}, EPLEX~\citep{la2019probabilistic}, and GP-GOMEA~\citep{virgolin2021improving} outperform the five baseline methods we choose from their study, in terms of $R^2$-based accuracy.
However, we find solution rate more aligned with edit distance, thus we choose the five best symbolic regression methods in terms of solution rate empirically shown for the \orgdataset datasets in SRBench~\citep{la2021contemporary}.
We also use three symbolic regression methods proposed in recent studies: a Transformer-based symbolic regression method referred to as E2E in~\citep{kamienny2022end}, unified deep symbolic regression (uDSR)~\citep{landajuela2022unified}, and a multi-population evolutionary algorithm named PySR~\citep{cranmer2023interpretable}.
For details of the baseline models, we refer readers to the corresponding papers~\citep{koza2005genetic,schmidt2011age,schmidt2009distilling,udrescu2020ai2,petersen2020deep,kamienny2022end,landajuela2022unified,cranmer2023interpretable}.
While we are aware that the research community is interested in performance of closed API services powered by language models such as ChatGPT~\footnote{\url{https://chat.openai.com/}}, it is not known what datasets are used to train the models behind the services.
We do not consider such services in this study since the benchmark uses our \ourproblem datasets on popular physics laws.\footnote{As of February 6th, 2024, we confirmed using our \ourproblem Easy set that ChatGPT-4 is not a strong baseline as given a \ourproblem dataset, it provides Python code to train linear regression models in scikit-learn or SR models in gplearn.}

\subsection{Runtime Constraints}
\label{subsec:runtime_constraints}
The implementations of the baseline methods in Section~\ref{subsec:baselines} except E2E\footnote{For E2E, we used an NVIDIA RTX 3090Ti.} do not use any GPUs.
We run 1,680 high performance computing (HPC) jobs in total, using compute nodes in an HPC cluster, which have 5 - 20 assigned physical CPU cores, 30 - 120 GB RAM, and 720 GB local storage.
Due to the properties of our resource, we have runtime constraints:
\begin{enumerate}[leftmargin=2em]
    \item Since each HPC job is designed to run for up to 24 hours due to the limited resource, we run a job with a pair of a target tabular dataset and a symbolic regression method.
    \item Given a pair of a dataset and a method, each of our HPC jobs runs up to 100 separate training sessions with different hyperparameter values (see Appendix~\ref{app_sec:baseline_hyperparams}).
\end{enumerate}

\subsection{Results}
\label{subsec:results}

We discuss experimental results of our baseline methods using \ourproblem datasets.
Tables~\ref{table:baseline_results} and~\ref{table:baseline_results_w_dummy_vars} summarize the performance of the baselines in terms of various metrics for the new datasets without/with dummy variables.
With our \ourproblem datasets, we confirm new findings and a different trend in the overall results compared to those in SRBench as summarized below:

\textbf{uDSR and PySR performed the best on our \ourproblem-Feynman datasets:}
According to $R^2$-based accuracy, uDSR significantly outperforms all the other baselines we considered for the \emph{Easy} and \emph{Medium} sets\footnote{Chi-squared tests for uDSR and PySR (the second best method in terms of $R^2$-based accuracy) showed p-values of $1.08 \times 10^{-4}$ and $1.37 \times 10^{-4}$, respectively.}, including AIF, which achieved the highest solution rate for the \orgdataset datasets in SRBench (see Appendix~\ref{app_sec:solution_rate_fsrd_vs_srsd}).
Note that $R^2$-based accuracy does not consider the interpretability of the prediction, which is a key property of SR methods and taken into account by solution rate and NED.
PySR produced more solutions structurally close to the true models than the other baseline methods and improved the other baseline methods in terms of solution rate and NED.
The results of uDSR and PySR also indicate difficulty levels of the three categories of our \ourproblem datasets, which looks aligned with our complexity-aware dataset categorization (Section~\ref{subsubsec:dataset_categories}).

\begin{table*}[t]
    \caption{Baseline results for \uline{\ourproblem-Feynman} from various perspectives: 1) accuracy ($R^2 > 0.999$)~\citep{la2021contemporary}, 2) solution rate~\citep{la2021contemporary}, and 3) NED (normalized edit distance).}
    \label{table:baseline_results}
    \vspace{-1em}
    \def\arraystretch{1.2}
    \small
    \begin{center}
    
        \bgroup
        \setlength{\tabcolsep}{0.4em}
        \begin{tabular}{c|l|rrrrrrrr} 
            \toprule
            \multirow{2}{*}{\bf Metric} & \multicolumn{1}{c|}{\multirow{2}{*}{\bf Group}} & \multicolumn{8}{c}{\bf \ourproblem-Feynman} \\
            \cline{3-10}
            & & \multicolumn{1}{c}{\bf gplearn} & \multicolumn{1}{c}{\bf AFP} & \multicolumn{1}{c}{\bf AFP-FE} & \multicolumn{1}{c}{\bf AIF} & \multicolumn{1}{c}{\bf DSR} & \multicolumn{1}{c}{\bf E2E} & \multicolumn{1}{c}{\bf uDSR} & \multicolumn{1}{c}{\bf PySR} \\
            \midrule
            \multirow{3}{*}{\rotatebox{90}{\shortstack{\small Accuracy\\{\scriptsize $R^2>0.999$}}}} & Easy & 6.67\% & 20.0\% & 26.7\% & 33.3\% & 63.3\% & 26.7\% & \highlight 100.0\% & 66.7\% \\ 
            & Medium & 7.50\% & 2.50\% & 2.50\% & 5.00\% & 45.0\% & 17.5\% & \highlight 75.0\% & 45.0\% \\ 
            & Hard & 2.00\% & 4.00\% & 4.00\% & 6.00\% & 28.0\% & 14.0\% & 20.0\% & \highlight 38.0\% \\
            \midrule
            \multirow{3}{*}{\rotatebox{90}{\shortstack{\small Solution \\\small Rate}}} & Easy & 6.67\% & 20.0\% & 23.3\% & 30.0\% & 46.7\% & 0.00\% & 50.0\% & \highlight 60.0\% \\ 
            & Medium & 0.00\% & 2.50\% & 2.50\% & 2.50\% & 10.0\% & 0.00\% & 17.5\% & \highlight 30.0\% \\ 
            & Hard & 0.00\% & 0.00\% & 0.00\% & 2.00\% & 2.00\% & 0.00\% & \highlight 4.00\% & \highlight 4.00\% \\
            \midrule
            \multirow{3}{*}{\rotatebox{90}{NED}} & Easy & 0.866 & 0.727 & 0.693 & 0.646 & 0.524 & 1.00 & 0.478 & \highlight 0.269 \\ 
            & Medium & 0.917 & 0.873 & 0.897 & 0.936 & 0.793 & 1.00 & 0.781 & \highlight 0.537 \\ 
            & Hard & 0.968 & 0.946 & 0.954 & 0.927 & 0.839 & 0.987 & 0.949 & \highlight 0.785 \\
            \bottomrule
        \end{tabular}
        \egroup
    \end{center}
    
    \caption{Baseline results for \uline{\ourproblem-Feynman with dummy variables} from various perspectives: 1) accuracy ($R^2 > 0.999$)~\citep{la2021contemporary}, 2) solution rate~\citep{la2021contemporary}, and 3) NED (normalized edit distance).}
    \label{table:baseline_results_w_dummy_vars}
    \vspace{-1em}
    \def\arraystretch{1.2}
    \small
    \begin{center}
        \bgroup
        \setlength{\tabcolsep}{0.4em}
        \begin{tabular}{c|l|rrrrrrrr} 
            \toprule
            \multirow{2}{*}{\bf Metric} & \multicolumn{1}{c|}{\multirow{2}{*}{\bf Group}} & \multicolumn{8}{c}{\bf \ourproblem-Feynman + Dummy Variables} \\
            \cline{3-10}
            & & \multicolumn{1}{c}{\bf gplearn} & \multicolumn{1}{c}{\bf AFP} & \multicolumn{1}{c}{\bf AFP-FE} & \multicolumn{1}{c}{\bf AIF} & \multicolumn{1}{c}{\bf DSR} & \multicolumn{1}{c}{\bf E2E} & \multicolumn{1}{c}{\bf uDSR} & \multicolumn{1}{c}{\bf PySR} \\
            \midrule
            \multirow{3}{*}{\rotatebox{90}{\shortstack{\small Accuracy\\{\scriptsize $R^2>0.999$}}}} & Easy & 0.00\% & 20.0\% & 16.7\% & 6.67\% & \highlight 76.7\% & 16.7\% & 53.3\% & 20.0\% \\ 
            & Medium & 0.00\% & 5.00\% & 0.00\% & 0.00\% & \highlight 45.0\% & 12.5\% & 37.5\% & 10.0\% \\ 
            & Hard & 0.00\% & 4.00\% & 4.00\% & 0.00\% & \highlight 22.0\% & 10.0\% & 12.0\% & 2.00\% \\
            \midrule
            \multirow{3}{*}{\rotatebox{90}{\shortstack{\small Solution \\\small Rate}}} & Easy & 0.00\% & 16.7\% & 16.7\% & 0.00\% & 10.0\% & 0.00\% & 10.0\% & \highlight 20.0\% \\ 
            & Medium & 0.00\% & 0.00\% & 0.00\% & 0.00\% & 0.00\% & 0.00\% & \highlight 7.50\% & 5.00\% \\ 
            & Hard & 0.00\% & 0.00\% & 0.00\% & 0.00\% & \highlight 2.00\% & 0.00\% & 0.00\% & 0.00\% \\
            \midrule
            \multirow{3}{*}{\rotatebox{90}{NED}} & Easy & 0.963 & 0.769 & 0.786 & 0.975 & 0.771 & 1.00 & 0.871 & \highlight 0.418 \\ 
            & Medium & 0.978 & 0.932 & 0.935 & 1.00 & 0.841 & 1.00 & 0.916 & \highlight 0.625 \\ 
            & Hard & 0.989 & 0.961 & 0.963 & 1.00 & \highlight 0.800 & 1.00 & 0.967 & 0.819 \\
            \bottomrule
        \end{tabular}
        \egroup
    \end{center}
\end{table*}
\begin{table*}[t]
    \caption{Percentages of predictions that use at least one dummy variable (\uline{left}) and those also considered ``correct'' as $R^2 > 0.999$ (\uline{right}). N/A: Denominator is zero.}
    \label{table:dummy_var_usage}
    \def\arraystretch{1.2}
    \begin{center}
        \bgroup
        \setlength{\tabcolsep}{0.2em}
        \begin{tabular}{l|rrrrrrrr}
            \toprule
            \multirow{2}{*}{\bf Group} & \multicolumn{8}{c}{\bf $\geq 1$ dummy variable used} \\
            \cline{2-9}
            & \multicolumn{1}{|c}{\bf gplearn} & \multicolumn{1}{c}{\bf AFP} & \multicolumn{1}{c}{\bf AFP-FE} & \multicolumn{1}{c}{\bf AIF} & \multicolumn{1}{c}{\bf DSR} & \multicolumn{1}{c}{\bf E2E} & \multicolumn{1}{c}{\bf uDSR} & \multicolumn{1}{c}{\bf PySR} \\
            \midrule
            Easy & 66.7\% & 52.9\% & 56.3\% & 50.0\% & 53.3\% & 100\% & 75.0\% & 63.3\% \\
            Medium & 0.00\% & 70.6\% & 43.8\% & \nacell & 59.0\% & 100\% & 66.7\% & 67.5\% \\
            Hard & 100.0\% & 81.8\% & 58.3\% & \nacell & 56.3\% & 100\% & 46.7\% & 64.0\% \\
        \end{tabular}        
        \begin{tabular}{l|rrrrrrrr}
            \toprule
            \multirow{2}{*}{\bf Group} & \multicolumn{8}{c}{\bf $\geq 1$ dummy variable used \& $R^2 > 0.999$} \\
            \cline{2-9}
            & \multicolumn{1}{|c}{\bf gplearn} & \multicolumn{1}{c}{\bf AFP} & \multicolumn{1}{c}{\bf AFP-FE} & \multicolumn{1}{c}{\bf AIF} & \multicolumn{1}{c}{\bf DSR} & \multicolumn{1}{c}{\bf E2E} & \multicolumn{1}{c}{\bf uDSR} & \multicolumn{1}{c}{\bf PySR} \\
            \midrule
            Easy & \nacell & 16.7\% & 0.00\% & 50.0\% & 47.8\% & 100\% & 75.0\% & 0.00\% \\
            Medium & \nacell & 50.0\% & \nacell & \nacell & 44.4\% & 100\% & 73.3\% & 0.00\% \\
            Hard & \nacell & 50.0\% & 0.00\% & \nacell & 36.4\% & 100\% & 33.3\% & 0.00\% \\
            \bottomrule
        \end{tabular}
        \egroup
    \end{center}
\end{table*}

\textbf{None of the baseline methods is robust against dummy variables:}
Overall, the performance differences between \ourproblem-Feynman datasets without and with dummy variables in Tables~\ref{table:baseline_results} and~\ref{table:baseline_results_w_dummy_vars} highlight that dummy variables made the \ourproblem problems even more challenging.
The dummy variable usage in Table~\ref{table:dummy_var_usage} indicates that all the baseline methods considered in this study failed to filter out random dummy variables, which the true models do not use and thus predicted solutions should not use.
It should be notable that while PySR's overall performance degraded due to the dummy variables, none of the PySR's solutions that include at least one dummy variable does not achieve $R^2 > 0.999$.
Even for \ourproblem datasets with dummy variables, PySR performed best among the considered baseline methods in terms of NED.
Those two trends suggest that it may be important in \ourproblem problems to penalize overcomplex solutions in a similar way to PySR~\citep{cranmer2023interpretable}.

\textbf{$R^2$-based accuracy is vulnerable to dummy variables:}
Table~\ref{table:baseline_results_w_dummy_vars} shows that compared to the results for \ourproblem-Feynman (Table~\ref{table:baseline_results}), AFP and DSR achieved comparable or even improved $R^2$-based accuracy for the datasets with dummy variables.
However, the results do not necessarily mean that those methods successfully filter out dummy variables.
While DSR performed incredibly better than other baseline methods in terms of $R^2$-based accuracy on the datasets with dummy variables, approximately 45.1\% of its non-zero predicted equations that meet $R^2 > 0.999$ (``correct'') use at least one dummy variable (Table~\ref{table:dummy_var_usage} (right)).
Similarly, 100\% of the ``correct'' equations from E2E meet the same conditions.

\textbf{NED provides a more fine-grained analysis than solution rate does:}
As pointed out in Section~\ref{subsec:metrics}, solution rate is based on binary evaluations$\textsuperscript{\ref{fn:solution_rate}}$ and does not help us how structurally close the predicted equation is to the true equation.
Thus, the solution rate may be not informative for challenging datasets, which is highlighted at the sparse rows of ``Solution Rate'' for \ourproblem-Feynman + Dummy Variables (especially Medium and Hard sets) in Table~\ref{table:baseline_results_w_dummy_vars}.
As demonstrated in the table, NED overcomes the drawback of solution rate and enables comparisons between AFP, AFP-FE, and DSR for the datasets with dummy variables while solution rate shows 0.00\% for most of the configurations.

\section{User Study: $R^2$ Score \& NED}
\label{sec:user_study}
To investigate how aligned with human judges the existing SR and new \ourproblem evaluation metrics are, we recruited 23 volunteers from industry and academia who either have doctoral degrees (scientists, professors, engineers) or are doctoral students, and performed a user study with approval from an ethics review board.
The volunteers are in diverse research fields such as computer science, mathematics, physics, chemistry, material science, aerospace engineering, engineering, medical nutrition, and computational biology.
Given a pair of true and estimate equations for an \ourproblem problem, the subjects were asked to assess an estimated equation on a discretized 1-to-5 scale, where 1 and 5 indicate ``1: Completely different from the true equation'' and ``5: Equivalent to the true equation'' respectively.
We chose \ourproblem problems among the 120 \ourproblem datasets such that we can obtain from the experimental results in Section~\ref{sec:experiments} at least two different equations estimated by different methods that are best among the baseline methods in terms of $R^2$ score and normalized edit distance (NED), respectively.
There were 24 resulting \ourproblem problems for the user study.
Note that we do not consider solution rate in the user study.
Different from $R^2$ score and NED (real values), we get a binary score (0 or 1) per predicted equation for solution rate, and the best predicted equation in terms of the binary score will be identical to the best predicted equation in terms of NED or randomly chosen from all the predicted equations if the best binary score is 0 \emph{i.e.}, none of the predicted equations is identical to the true equation.

\begin{wraptable}{r}{20em}
    \vspace{-1em}
    \caption{Pearson correlation coefficients (PCCs) between the human judges and SR/\ourproblem metrics.}
    \vspace{-1.5em}
    \label{table:user_study}
    \small
    \begin{center}
        \begin{tabular}{l|rr}
            \toprule
            \multicolumn{1}{c|}{\bf Metrics} & \multicolumn{1}{c}{\bf PCC} & \multicolumn{1}{c}{\bf P-value} \\
            \midrule
            SR: $R^2$ score & $4.66 \times 10^{-3}$ & $0.913$ \\
            \ourproblem: NED & \highlight $-0.416$ & \highlight $1.85 \times 10^{-24}$\\
            \bottomrule
        \end{tabular}
    \end{center}
\end{wraptable}

Table~\ref{table:user_study} shows Pearson correlation coefficients (PCCs) between the human judges and SR/\ourproblem evaluation metrics.
For normalized edit distance (NED), the Pearson correlation coefficient and p-value were $-0.416$ and $1.85 \times 10^{-24}$ respectively, which show a much stronger and statistically more significant correlation between NED and human judges than one for $R^2$ scores.
In other words, the results suggest that normalized edit distance is more aligned with human judges than $R^2$ score, and thus can be a better estimate of how close to the true equations the estimated equations are, in a more human-understandable way.
Note that the smaller an NED is, the better (structurally closer to the true model) the solution is, thus the negative correlation coefficient is expected.

\section{Conclusion}
\label{sec:conclusion}
In this work, we pointed out issues of existing datasets (\emph{e.g.}, \orgdataset) and benchmarks of symbolic regression for scientific discovery (\ourproblem).
To address the issues, we proposed 1) 120 new \ourproblem datasets based on a set of physics formulas in \orgdataset~\citep{udrescu2020ai}, 2) another 120 new \ourproblem datasets containing dummy variables, and 3) a new evaluation metric for \ourproblem to discuss the structural similarity between the true and estimated symbolic expressions (equations).
The benchmark results revealed key findings including uDSR and PySR being the state of the art for the \ourproblem datasets (AIF performed the best for the ~\orgdataset datasets in SRBench~\citep{la2021contemporary}) and the vulnerability of both the SR baselines and $R^2$-based accuracy to dummy variables in \ourproblem problems.
The experimental results and user study also suggest that the normalized edit distance is an additional reasonable metric for \ourproblem, which provides a more fine-grained analysis than solution rate and incorporates existing SR metrics (\emph{e.g.}, Tables~\ref{table:baseline_results} and~\ref{table:baseline_results_w_dummy_vars}).
We also summarize the limitations of this work in Appendix~\ref{app_sec:limitations}.
Last but not least, the experimental results suggest that our \ourproblem datasets are more challenging for existing SR methods than the \orgdataset datasets, and there is still significant room for improvement in SR methods for \ourproblem problems according to the metrics considered in this study.
To encourage the studies of \ourproblem, we publish our datasets and code with Creative Commons Attribution 4.0 (CC BY) and MIT License, respectively.

\impact{Intended uses of our proposed datasets and evaluation criteria are for scientific knowledge discoveries such as hidden laws in physics. The datasets consist of physics formulas and do not include any social/personal information. We believe that a potential positive societal consequence of this work is that our benchmark datasets and evaluation criteria will help non-ML experts choose which symbolic regression (SR) methods they want to apply to their problem for scientific discoveries. The proposed datasets and evaluation criteria are more realistic for discussing scientific discovery than existing SR datasets and evaluation criteria as 1) we carefully reviewed the properties of the physics formulas and designed new annotation policies for the proposed datasets, 2) we introduced dummy variables to the proposed datasets to discuss the robustness of SR methods, and 3) our user study shows that the proposed NED is more aligned with human judges than a popular existing SR metric. There are also important considerations regarding the benchmark. When applying SR method to real-world problems, there must be observation noises. However, simulating such realistic noise injections for creating datasets is a challenging task, and our benchmark might bias research in favor of methods that work well for problems where such realistic noises do not exist. Instead, we introduced an assumption that the observed data may include dummy variables, which are not necessary to explain a hidden law.}

\acks{This work was supported by JST-Mirai Program Grant Number JPMJMI21G2, JST Moonshot R\&D Program Grant Number JPMJMS2236, and JSPS KAKENHI Grant Number 21K14130, Japan.
We used computational resource of AI Bridging Cloud Infrastructure (ABCI) provided by National Institute of Advanced Industrial Science and Technology (AIST).}

\vskip 0.2in
\bibliography{references}

\begin{thebibliography}{54}
\providecommand{\natexlab}[1]{#1}
\providecommand{\url}[1]{\texttt{#1}}
\expandafter\ifx\csname urlstyle\endcsname\relax
  \providecommand{\doi}[1]{doi: #1}\else
  \providecommand{\doi}{doi: \begingroup \urlstyle{rm}\Url}\fi

\bibitem[Abdellaoui and Mehrkanoon(2021)]{abdellaoui2021symbolic}
Ismail~Alaoui Abdellaoui and Siamak Mehrkanoon.
\newblock Symbolic regression for scientific discovery: an application to wind speed forecasting.
\newblock In \emph{2021 IEEE Symposium Series on Computational Intelligence (SSCI)}, pages 01--08. IEEE, 2021.

\bibitem[Akiba et~al.(2019)Akiba, Sano, Yanase, Ohta, and Koyama]{akiba2019optuna}
Takuya Akiba, Shotaro Sano, Toshihiko Yanase, Takeru Ohta, and Masanori Koyama.
\newblock {Optuna: A Next-generation Hyperparameter Optimization Framework}.
\newblock In \emph{Proceedings of the 25th ACM SIGKDD international conference on knowledge discovery \& data mining}, pages 2623--2631, 2019.

\bibitem[Batra et~al.(2020)Batra, Song, and Ramprasad]{batra2020emerging}
Rohit Batra, Le~Song, and Rampi Ramprasad.
\newblock Emerging materials intelligence ecosystems propelled by machine learning.
\newblock \emph{Nature Reviews Materials}, pages 1--24, 2020.

\bibitem[Biggio et~al.(2021)Biggio, Bendinelli, Neitz, Lucchi, and Parascandolo]{biggio2021neural}
Luca Biggio, Tommaso Bendinelli, Alexander Neitz, Aurelien Lucchi, and Giambattista Parascandolo.
\newblock {Neural Symbolic Regression that Scales}.
\newblock In \emph{International Conference on Machine Learning}, pages 936--945. PMLR, 2021.

\bibitem[Burke et~al.(2002)Burke, Gustafson, Kendall, and Krasnogor]{burke2002advanced}
Edmund Burke, Steven Gustafson, Graham Kendall, and Natalio Krasnogor.
\newblock {Advanced Population Diversity Measures in Genetic Programming}.
\newblock In \emph{International Conference on Parallel Problem Solving from Nature}, pages 341--350. Springer, 2002.

\bibitem[Cranmer(2023)]{cranmer2023interpretable}
Miles Cranmer.
\newblock {Interpretable Machine Learning for Science with PySR and SymbolicRegression.jl}.
\newblock \emph{arXiv preprint arXiv:2305.01582}, 2023.

\bibitem[Cranmer et~al.(2020)Cranmer, Sanchez~Gonzalez, Battaglia, Xu, Cranmer, Spergel, and Ho]{cranmer2020discovering}
Miles Cranmer, Alvaro Sanchez~Gonzalez, Peter Battaglia, Rui Xu, Kyle Cranmer, David Spergel, and Shirley Ho.
\newblock {Discovering Symbolic Models from Deep Learning with Inductive Biases}.
\newblock \emph{Advances in Neural Information Processing Systems}, 33:\penalty0 17429--17442, 2020.

\bibitem[Feynman et~al.(1963{\natexlab{a}})Feynman, Leighton, and Sands]{feynman1963feynman1}
Richard~P Feynman, Robert~B Leighton, and Matthew Sands.
\newblock \emph{{The Feynman Lectures on Physics, Vol. I: The New Millennium Edition: Mainly Mechanics, Radiation, and Heat}}, volume~1.
\newblock Basic books, 1963{\natexlab{a}}.

\bibitem[Feynman et~al.(1963{\natexlab{b}})Feynman, Leighton, and Sands]{feynman1963feynman2}
Richard~P Feynman, Robert~B Leighton, and Matthew Sands.
\newblock \emph{{The Feynman Lectures on Physics, Vol. II: The New Millennium Edition: Mainly Electromagnetism and Matter}}, volume~2.
\newblock Basic books, 1963{\natexlab{b}}.

\bibitem[Feynman et~al.(1963{\natexlab{c}})Feynman, Leighton, and Sands]{feynman1963feynman3}
Richard~P Feynman, Robert~B Leighton, and Matthew Sands.
\newblock \emph{{The Feynman Lectures on Physics, Vol. II: The New Millennium Edition: Quantum Mechanics}}, volume~3.
\newblock Basic books, 1963{\natexlab{c}}.

\bibitem[Goldstein et~al.(2002)Goldstein, Poole, and Safko]{goldstein2002classical}
Herbert Goldstein, Charles Poole, and John Safko.
\newblock \emph{{Classical Mechanics}}.
\newblock Addison Wesley, 2002.

\bibitem[Hoai et~al.(2002)Hoai, McKay, Essam, and Chau]{hoai2002solving}
Nguyen~Xuan Hoai, Robert~Ian McKay, D~Essam, and R~Chau.
\newblock {Solving the Symbolic Regression Problem with Tree Adjunct Grammar Guided Genetic Programming: The Comparative Results}.
\newblock In \emph{Proceedings of the 2002 Congress on Evolutionary Computation. CEC'02 (Cat. No. 02TH8600)}, volume~2, pages 1326--1331. IEEE, 2002.

\bibitem[Huang et~al.(2021)Huang, Li, Huang, Wang, and Jiang]{huang2021ai}
Zhanchao Huang, Chunjiang Li, Zhilong Huang, Yong Wang, and Hanqing Jiang.
\newblock {AI-Timoshenko: Automatedly Discovering Simplified Governing Equations for Applied Mechanics Problems From Simulated Data}.
\newblock \emph{Journal of Applied Mechanics}, 88\penalty0 (10):\penalty0 101006, 2021.

\bibitem[Jackson(1999)]{jackson1999classical}
John~David Jackson.
\newblock \emph{{Classical Electrodynamics}}.
\newblock Wiley, 1999.

\bibitem[Jin et~al.(2019)Jin, Fu, Kang, Guo, and Guo]{jin2019bayesian}
Ying Jin, Weilin Fu, Jian Kang, Jiadong Guo, and Jian Guo.
\newblock {Bayesian Symbolic Regression}.
\newblock \emph{arXiv preprint arXiv:1910.08892}, 2019.

\bibitem[Johnson(2009)]{johnson2009genetic}
Colin~G Johnson.
\newblock {Genetic Programming Crossover: Does it Cross Over?}
\newblock In \emph{European Conference on Genetic Programming}, pages 97--108. Springer, 2009.

\bibitem[Kamienny et~al.(2022)Kamienny, d'Ascoli, Lample, and Charton]{kamienny2022end}
Pierre-Alexandre Kamienny, St{\'e}phane d'Ascoli, Guillaume Lample, and Fran{\c{c}}ois Charton.
\newblock {End-to-end Symbolic Regression with Transformers}.
\newblock In \emph{Advances in Neural Information Processing Systems}, volume~35, pages 10269--10281, 2022.

\bibitem[Keijzer(2003)]{keijzer2003improving}
Maarten Keijzer.
\newblock {Improving Symbolic Regression with Interval Arithmetic and Linear Scaling}.
\newblock In \emph{European Conference on Genetic Programming}, pages 70--82. Springer, 2003.

\bibitem[Kim et~al.(2020)Kim, Lu, Mukherjee, Gilbert, Jing, {\v{C}}eperi{\'c}, and Solja{\v{c}}i{\'c}]{kim2020integration}
Samuel Kim, Peter~Y Lu, Srijon Mukherjee, Michael Gilbert, Li~Jing, Vladimir {\v{C}}eperi{\'c}, and Marin Solja{\v{c}}i{\'c}.
\newblock {Integration of Neural Network-Based Symbolic Regression in Deep Learning for Scientific Discovery}.
\newblock \emph{IEEE Transactions on Neural Networks and Learning Systems}, 2020.

\bibitem[Kommenda et~al.(2020)Kommenda, Burlacu, Kronberger, and Affenzeller]{kommenda2020parameter}
Michael Kommenda, Bogdan Burlacu, Gabriel Kronberger, and Michael Affenzeller.
\newblock {Parameter identification for symbolic regression using nonlinear least squares}.
\newblock \emph{Genetic Programming and Evolvable Machines}, 21\penalty0 (3):\penalty0 471--501, 2020.

\bibitem[Koza and Poli(2005)]{koza2005genetic}
John~R Koza and Riccardo Poli.
\newblock {Genetic Programming}.
\newblock In \emph{Search methodologies}, pages 127--164. Springer, 2005.

\bibitem[La~Cava et~al.(2016)La~Cava, Danai, and Spector]{la2016inference}
William La~Cava, Kourosh Danai, and Lee Spector.
\newblock {Inference of compact nonlinear dynamic models by epigenetic local search}.
\newblock \emph{Engineering Applications of Artificial Intelligence}, 55:\penalty0 292--306, 2016.

\bibitem[La~Cava et~al.(2018)La~Cava, Singh, Taggart, Suri, and Moore]{la2018learning}
William La~Cava, Tilak~Raj Singh, James Taggart, Srinivas Suri, and Jason~H Moore.
\newblock {Learning concise representations for regression by evolving networks of trees}.
\newblock In \emph{International Conference on Learning Representations}, 2018.

\bibitem[La~Cava et~al.(2019)La~Cava, Helmuth, Spector, and Moore]{la2019probabilistic}
William La~Cava, Thomas Helmuth, Lee Spector, and Jason~H Moore.
\newblock {A Probabilistic and Multi-Objective Analysis of Lexicase Selection and $\varepsilon$-Lexicase Selection}.
\newblock \emph{Evolutionary Computation}, 27\penalty0 (3):\penalty0 377--402, 2019.

\bibitem[La~Cava et~al.(2021)La~Cava, Orzechowski, Burlacu, de~Franca, Virgolin, Jin, Kommenda, and Moore]{la2021contemporary}
William La~Cava, Patryk Orzechowski, Bogdan Burlacu, Fabricio~Olivetti de~Franca, Marco Virgolin, Ying Jin, Michael Kommenda, and Jason~H Moore.
\newblock {Contemporary Symbolic Regression Methods and their Relative Performance}.
\newblock In \emph{Thirty-fifth Conference on Neural Information Processing Systems Datasets and Benchmarks Track (Round 1)}, 2021.

\bibitem[Landajuela et~al.(2021)Landajuela, Petersen, Kim, Santiago, Glatt, Mundhenk, Pettit, and Faissol]{landajuela2021discovering}
Mikel Landajuela, Brenden~K Petersen, Sookyung Kim, Claudio~P Santiago, Ruben Glatt, Nathan Mundhenk, Jacob~F Pettit, and Daniel Faissol.
\newblock Discovering symbolic policies with deep reinforcement learning.
\newblock In \emph{International Conference on Machine Learning}, pages 5979--5989. PMLR, 2021.

\bibitem[Landajuela et~al.(2022)Landajuela, Lee, Yang, Glatt, Santiago, Aravena, Mundhenk, Mulcahy, and Petersen]{landajuela2022unified}
Mikel Landajuela, Chak~Shing Lee, Jiachen Yang, Ruben Glatt, Claudio~P Santiago, Ignacio Aravena, Terrell Mundhenk, Garrett Mulcahy, and Brenden~K Petersen.
\newblock {A Unified Framework for Deep Symbolic Regression}.
\newblock \emph{Advances in Neural Information Processing Systems}, 35:\penalty0 33985--33998, 2022.

\bibitem[Liu et~al.(2021)Liu, Wang, Meng, Chen, Tegmark, and Liu]{liu2021nnphd}
Z~Liu, B~Wang, Q~Meng, W~Chen, M~Tegmark, and TY~Liu.
\newblock Machine-learning nonconservative dynamics for new-physics detection.
\newblock \emph{Physical Review. E}, 104\penalty0 (5-2):\penalty0 055302--055302, 2021.

\bibitem[Liu and Tegmark(2021)]{liu2021aipoincare}
Ziming Liu and Max Tegmark.
\newblock {Machine Learning Conservation Laws from Trajectories}.
\newblock \emph{Physical Review Letters}, 126:\penalty0 180604, May 2021.

\bibitem[Loftis et~al.(2020)Loftis, Yuan, Zhao, Hu, and Hu]{loftis2020lattice}
Christian Loftis, Kunpeng Yuan, Yong Zhao, Ming Hu, and Jianjun Hu.
\newblock {Lattice Thermal Conductivity Prediction Using Symbolic Regression and Machine Learning}.
\newblock \emph{The Journal of Physical Chemistry A}, 125\penalty0 (1):\penalty0 435--450, 2020.

\bibitem[Meurer et~al.(2017)Meurer, Smith, Paprocki, {\v{C}}ert{\'\i}k, Kirpichev, Rocklin, Kumar, Ivanov, Moore, Singh, et~al.]{meurer2017sympy}
Aaron Meurer, Christopher~P Smith, Mateusz Paprocki, Ond{\v{r}}ej {\v{C}}ert{\'\i}k, Sergey~B Kirpichev, Matthew Rocklin, AMiT Kumar, Sergiu Ivanov, Jason~K Moore, Sartaj Singh, et~al.
\newblock {SymPy: symbolic computing in Python}.
\newblock \emph{PeerJ Computer Science}, 3:\penalty0 e103, 2017.

\bibitem[Nakai et~al.(2013)Nakai, Miyahara, Kuboyama, Uchida, and Suzuki]{nakai2013acquisition}
Shohei Nakai, Tetsuhiro Miyahara, Tetsuji Kuboyama, Tomoyuki Uchida, and Yusuke Suzuki.
\newblock {Acquisition of Characteristic Tree Patterns with VLDC's by Genetic Programming and Edit Distance}.
\newblock In \emph{2013 Second IIAI International Conference on Advanced Applied Informatics}, pages 147--151. IEEE, 2013.

\bibitem[{National Astronomical Observatory of Japan}(2022)]{nationalhandbook2022}
{National Astronomical Observatory of Japan}.
\newblock \emph{{Handbook of Scientific Tables}}.
\newblock World Scientific, 2022.

\bibitem[O'Reilly(1997)]{o1997using}
U-M O'Reilly.
\newblock {Using a distance metric on genetic programs to understand genetic operators}.
\newblock In \emph{1997 IEEE International Conference on Systems, Man, and Cybernetics. Computational Cybernetics and Simulation}, volume~5, pages 4092--4097. IEEE, 1997.

\bibitem[Orzechowski et~al.(2018)Orzechowski, La~Cava, and Moore]{orzechowski2018we}
Patryk Orzechowski, William La~Cava, and Jason~H Moore.
\newblock {Where are we now? A large benchmark study of recent symbolic regression methods}.
\newblock In \emph{Proceedings of the Genetic and Evolutionary Computation Conference}, pages 1183--1190, 2018.

\bibitem[Petersen et~al.(2020)Petersen, Larma, Mundhenk, Santiago, Kim, and Kim]{petersen2020deep}
Brenden~K Petersen, Mikel~Landajuela Larma, Terrell~N Mundhenk, Claudio~Prata Santiago, Soo~Kyung Kim, and Joanne~Taery Kim.
\newblock {Deep symbolic regression: Recovering mathematical expressions from data via risk-seeking policy gradients}.
\newblock In \emph{International Conference on Learning Representations}, 2020.

\bibitem[Przybocki et~al.(2006)Przybocki, Sanders, and Le]{przybocki2006edit}
Mark Przybocki, Gregory Sanders, and Audrey Le.
\newblock {Edit Distance: A Metric for Machine Translation Evaluation}.
\newblock In \emph{Proceedings of the Fifth International Conference on Language Resources and Evaluation (LREC’06)}, 2006.

\bibitem[Schmidt and Lipson(2009)]{schmidt2009distilling}
Michael Schmidt and Hod Lipson.
\newblock {Distilling Free-Form Natural Laws from Experimental Data}.
\newblock \emph{Science}, 324\penalty0 (5923):\penalty0 81--85, 2009.

\bibitem[Schmidt and Lipson(2011)]{schmidt2011age}
Michael Schmidt and Hod Lipson.
\newblock {Age-Fitness Pareto Optimization}.
\newblock In \emph{Genetic programming theory and practice VIII}, pages 129--146. Springer, 2011.

\bibitem[Schwartz(2014)]{schwartz2014quantum}
Matthew~D Schwartz.
\newblock \emph{{Quantum Field Theory and the Standard Model}}.
\newblock Cambridge University Press, 2014.

\bibitem[Strogatz(2018)]{strogatz2018nonlinear}
Steven~H Strogatz.
\newblock \emph{{Nonlinear Dynamics and Chaos with Student Solutions Manual: With Applications to Physics, Biology, Chemistry, and Engineering}}.
\newblock CRC press, 2018.

\bibitem[Sun et~al.(2019)Sun, Ouyang, Zhang, and Zhang]{sun2019data}
Sheng Sun, Runhai Ouyang, Bochao Zhang, and Tong-Yi Zhang.
\newblock Data-driven discovery of formulas by symbolic regression.
\newblock \emph{MRS Bulletin}, 44\penalty0 (7):\penalty0 559--564, 2019.

\bibitem[Trujillo et~al.(2016)Trujillo, Mu{\~n}oz, Galv{\'a}n-L{\'o}pez, and Silva]{trujillo2016neat}
Leonardo Trujillo, Luis Mu{\~n}oz, Edgar Galv{\'a}n-L{\'o}pez, and Sara Silva.
\newblock {neat Genetic Programming: Controlling bloat naturally}.
\newblock \emph{Information Sciences}, 333:\penalty0 21--43, 2016.

\bibitem[Udrescu and Tegmark(2020)]{udrescu2020ai}
Silviu-Marian Udrescu and Max Tegmark.
\newblock {AI Feynman: A physics-inspired method for symbolic regression}.
\newblock \emph{Science Advances}, 6\penalty0 (16):\penalty0 eaay2631, 2020.

\bibitem[Udrescu et~al.(2020)Udrescu, Tan, Feng, Neto, Wu, and Tegmark]{udrescu2020ai2}
Silviu-Marian Udrescu, Andrew Tan, Jiahai Feng, Orisvaldo Neto, Tailin Wu, and Max Tegmark.
\newblock {AI Feynman 2.0: Pareto-optimal symbolic regression exploiting graph modularity}.
\newblock \emph{Advances in Neural Information Processing Systems}, 33, 2020.

\bibitem[Uy et~al.(2011)Uy, Hoai, O’Neill, McKay, and Galv{\'a}n-L{\'o}pez]{uy2011semantically}
Nguyen~Quang Uy, Nguyen~Xuan Hoai, Michael O’Neill, Robert~I McKay, and Edgar Galv{\'a}n-L{\'o}pez.
\newblock {Semantically-based Crossover in Genetic Programming: Application to Real-valued Symbolic Regression}.
\newblock \emph{Genetic Programming and Evolvable Machines}, 12\penalty0 (2):\penalty0 91--119, 2011.

\bibitem[Valipour et~al.(2021)Valipour, You, Panju, and Ghodsi]{valipour2021symbolicgpt}
Mojtaba Valipour, Bowen You, Maysum Panju, and Ali Ghodsi.
\newblock {SymbolicGPT: A Generative Transformer Model for Symbolic Regression}.
\newblock \emph{arXiv preprint arXiv:2106.14131}, 2021.

\bibitem[Virgolin et~al.(2019)Virgolin, Alderliesten, and Bosman]{virgolin2019linear}
Marco Virgolin, Tanja Alderliesten, and Peter~AN Bosman.
\newblock {Linear scaling with and within semantic backpropagation-based genetic programming for symbolic regression}.
\newblock In \emph{Proceedings of the genetic and evolutionary computation conference}, pages 1084--1092, 2019.

\bibitem[Virgolin et~al.(2021)Virgolin, Alderliesten, Witteveen, and Bosman]{virgolin2021improving}
Marco Virgolin, Tanja Alderliesten, Cees Witteveen, and Peter~AN Bosman.
\newblock {Improving Model-Based Genetic Programming for Symbolic Regression of Small Expressions}.
\newblock \emph{Evolutionary computation}, 29\penalty0 (2):\penalty0 211--237, 2021.

\bibitem[Wang et~al.(2019)Wang, Wagner, and Rondinelli]{wang2019symbolic}
Yiqun Wang, Nicholas Wagner, and James~M Rondinelli.
\newblock Symbolic regression in materials science.
\newblock \emph{MRS Communications}, 9\penalty0 (3):\penalty0 793--805, 2019.

\bibitem[Weinberg(1972)]{weinberg1972gravitation}
Steven Weinberg.
\newblock \emph{{Gravitation and Cosmology: Principles and Applications of the General Theory of Relativity}}.
\newblock Wiley,, 1972.

\bibitem[Weng et~al.(2020)Weng, Song, Zhu, Yan, Sun, Grice, Yan, and Yin]{weng2020simple}
Baicheng Weng, Zhilong Song, Rilong Zhu, Qingyu Yan, Qingde Sun, Corey~G Grice, Yanfa Yan, and Wan-Jian Yin.
\newblock Simple descriptor derived from symbolic regression accelerating the discovery of new perovskite catalysts.
\newblock \emph{Nature Communications}, 11\penalty0 (1):\penalty0 1--8, 2020.

\bibitem[Wu and Tegmark(2019)]{wu2019toward}
Tailin Wu and Max Tegmark.
\newblock Toward an artificial intelligence physicist for unsupervised learning.
\newblock \emph{Physical Review E}, 100\penalty0 (3):\penalty0 033311, 2019.

\bibitem[Zhang and Shasha(1989)]{zhang1989simple}
Kaizhong Zhang and Dennis Shasha.
\newblock {Simple Fast Algorithms for the Editing Distance between Trees and Related Problems}.
\newblock \emph{SIAM journal on computing}, 18\penalty0 (6):\penalty0 1245--1262, 1989.

\end{thebibliography}

\appendix

\section{Our \ourproblem Datasets: Additional Information}
\label{app_sec:dataset_additional_info}
This section provides additional information about our \ourproblem datasets.
We created the datasets to discuss the performance of symbolic regression for scientific discovery (\ourproblem).
Each of the 120 \ourproblem datasets consists of 10,000 samples and has train, val, and test splits with ratio of 8:1:1.
For the annotation policy of our \ourproblem datasets, we refer readers to Section~\ref{subsubsec:annotation_policy}.

Tables~\ref{table:easy2} --~\ref{table:hard5} comprehensively summarize the differences between \orgdataset and our \ourproblem datasets.
Note that the table of Easy set (part 1) is provided as Table~\ref{table:easy1} in Section~\ref{subsec:dataset_issues}.
As described in Section~\ref{subsubsec:dataset_categories}, we categorized each of the 120 \ourproblem datasets into one of Easy, Medium, and Hard sets.
We also introduced dummy variables to the three groups of the datasets, which created another three groups of the datasets (see Section~\ref{subsec:dummy_variables}).
We published the six groups of the \ourproblem datasets with Creative Commons Attribution 4.0 and DOIs (digital object identifiers) at Hugging Face Dataset repositories.%
The dataset documentations are publicly available as Hugging Face Dataset cards, where we additionally provide an SI derived unit and an SI unit for each of all the constants and variables due to the limited horizontal space in this paper.
We also published our codebase as a GitHub repository.\textsuperscript{\ref{fn:code_repo}}
These repositories are version-controlled with Git~\footnote{\url{https://git-scm.com/}} so that users can track the log of the changes.
We bear all responsibility in case of violation of rights.

\begin{table}[H]
    \caption{\uline{Easy set} of our proposed datasets (part 2). C: Constant, V: Variable, F: Float, I: Integer, P: Positive, N: Negative, NN: Non-Negative, I$\star$: Integer treated as float due to the capacity of 32-bit integer, $\U$: Uniform distribution, $\Ulog$: Log-Uniform distribution.}
    \label{table:easy2}
    \def\arraystretch{1.2}
    \begin{center}
        \bgroup
        \setlength{\tabcolsep}{0.1em}
        \scriptsize
\begin{tabular}{lcllcccc} 
            \toprule
            \multicolumn{1}{c}{\multirow{2}{*}{\bf Eq. ID}} & \multicolumn{1}{c}{\multirow{2}{*}{\bf Formula}} & \multicolumn{2}{c}{\multirow{2}{*}{\bf Symbols}} & \multicolumn{2}{c}{\bf Properties} & \multicolumn{2}{c}{\bf Distributions}
            \\ %
            & & & & Original & Ours & Original & Ours \\\midrule
            \multirow{4}{*}{I.30.5} & \multirow{4}{*}{$d = \frac{\lambda}{n \sin\theta}$} & $d$ & Interplanar distance & V, F, P & V, F, P & $\U(2, 5)$ & N/A\\
             &  & $\lambda$ & Wavelength of X-ray & V, F, P & V, F, P & $\U(1, 2)$ & $\Ulog(10^{-11}, 10^{-9})$\\
             &  & $n$ & Number of phase difference & V, F, P & V, I, P & $\U(1, 5)$ & $\Ulog(10^{0}, 10^{2})$\\
             &  & $\theta$ & Incidence/Reflection angle & V, F, P & V, F, NN & N/A & $\U(0, \frac{\pi}{2})$\\\hline
            \multirow{5}{*}{I.43.16} & \multirow{5}{*}{$v = \mu q \frac{V}{d}$} & $v$ & Velocity & V, F, P & V, F & N/A & N/A\\
             &  & $\mu$ & Ionic conductivity & V, F, P & V, F & $\U(1, 5)$ & $\Ulog(10^{-6}, 10^{-4})$\\
\bottomrule
        \end{tabular}
        \egroup
    \end{center}
\end{table}

\begin{table}[H]
    \caption{\uline{Easy set} of our proposed datasets (part 3).}
    \label{table:easy3}
    \def\arraystretch{1.2}
    \begin{center}
        \bgroup
        \setlength{\tabcolsep}{0.25em}
        \scriptsize
        \begin{tabular}{lcllcccc} 
            \toprule
            \multicolumn{1}{c}{\multirow{2}{*}{\bf Eq. ID}} & \multicolumn{1}{c}{\multirow{2}{*}{\bf Formula}} & \multicolumn{2}{c}{\multirow{2}{*}{\bf Symbols}} & \multicolumn{2}{c}{\bf Properties} & \multicolumn{2}{c}{\bf Distributions}
            \\ %
            & & & & Original & Ours & Original & Ours \\\midrule
            \multirow{4}{*}{II.15.5} & \multirow{4}{*}{$U = -p E \cos\theta$} & $U$ & Energy & V, F & V, F & N/A & N/A\\
             &  & $p$ & Electric dipole moment & V, F, P & V, F & $\U(1, 5)$ & $\Ulog(10^{-22}, 10^{-20})$\\
             &  & $E$ & Magnitude of electric field & V, F, P & V, F & $\U(1, 5)$ & $\Ulog(10^{1}, 10^{3})$\\
             &  & $\theta$ & Angle & V, F, P & V, F, NN & $\U(1, 5)$ & $\U(0, 2 \pi)$\\\hline
            \multirow{4}{*}{II.27.16} & \multirow{4}{*}{$L = \epsilon c E^2$} & $L$ & Radiance & V, F, P & V, F & N/A & N/A\\
             &  & $\epsilon$ & Vacuum permittivity & V, F, P & C, F, P & $\U(1, 5)$ & $8.854 \times 10^{-12}$\\
\bottomrule
        \end{tabular}
        \egroup
    \end{center}
\end{table}

\begin{table}[H]
    \caption{\uline{Medium set} of our proposed datasets (part 1).}
    \label{table:medium1}
    \def\arraystretch{1.15}
    \begin{center}
        \bgroup
        \setlength{\tabcolsep}{0.2em}
        \scriptsize
        \begin{tabular}{lcllcccc} 
            \toprule
            \multicolumn{1}{c}{\multirow{2}{*}{\bf Eq. ID}} & \multicolumn{1}{c}{\multirow{2}{*}{\bf Formula}} & \multicolumn{2}{c}{\multirow{2}{*}{\bf Symbols}} & \multicolumn{2}{c}{\bf Properties} & \multicolumn{2}{c}{\bf Distributions}
            \\ %
            & & & & Original & Ours & Original & Ours \\\midrule
            \multirow{5}{*}{I.8.14} & \multirow{5}{*}{$d = \sqrt{(x_2-x_1)^2+(y_2-y_1)^2}$} & $d$ & Distance & V, F, P & V, F, NN & N/A & N/A\\
             &  & $x_2$ & Position & V, F, P & V, F & $\U(1, 5)$ & $\Ulog(10^{-1}, 10^{1})$\\
             &  & $x_1$ & Position & V, F, P & V, F & $\U(1, 5)$ & $\Ulog(10^{-1}, 10^{1})$\\
             &  & $y_2$ & Position & V, F, P & V, F & $\U(1, 5)$ & $\Ulog(10^{-1}, 10^{1})$\\
             &  & $y_1$ & Position & V, F, P & V, F & $\U(1, 5)$ & $\Ulog(10^{-1}, 10^{1})$\\\hline
            \multirow{4}{*}{I.10.7} & \multirow{4}{*}{$m = \frac{m_0}{\sqrt{1-\frac{v^2}{c^2}}}$} & $m$ & Mass & V, F, P & V, F, P & N/A & N/A\\
\bottomrule
        \end{tabular}
        \egroup
    \end{center}
\end{table}

\begin{table}[H]
    \caption{\uline{Medium set} of our proposed datasets (part 2).}
    \label{table:medium2}
    \def\arraystretch{1.2}
    \begin{center}
        \bgroup
        \setlength{\tabcolsep}{0.15em}
        \scriptsize
        \begin{tabular}{lcllcccc} 
            \toprule
            \multicolumn{1}{c}{\multirow{2}{*}{\bf Eq. ID}} & \multicolumn{1}{c}{\multirow{2}{*}{\bf Formula}} & \multicolumn{2}{c}{\multirow{2}{*}{\bf Symbols}} & \multicolumn{2}{c}{\bf Properties} & \multicolumn{2}{c}{\bf Distributions}
            \\ %
            & & & & Original & Ours & Original & Ours \\\midrule
            \multirow{5}{*}{I.24.6} & \multirow{5}{*}{$E = \frac{1}{4} m (\omega^2+\omega_0^2) x^2$} & $E$ & Energy & V, F, P & V, F, P & N/A & N/A\\
             &  & $m$ & Mass & V, F, P & V, F, P & $\U(1, 3)$ & $\Ulog(10^{-1}, 10^{1})$\\
             &  & $\omega$ & Angular velocity & V, F, P & V, F & $\U(1, 3)$ & $\Ulog(10^{-1}, 10^{1})$\\
             &  & $\omega_0$ & Angular velocity & V, F, P & V, F & $\U(1, 3)$ & $\Ulog(10^{-1}, 10^{1})$\\
             &  & $x$ & Position & V, F, P & V, F & $\U(1, 3)$ & $\Ulog(10^{-1}, 10^{1})$\\\hline
            \multirow{3}{*}{I.29.4} & \multirow{3}{*}{$k = \frac{\omega}{c}$} & $k$ & Wavenumber & V, F, P & V, F, P & N/A & N/A\\
\bottomrule
        \end{tabular}
        \egroup
    \end{center}
\end{table}

\begin{table}[H]
    \caption{\uline{Medium set} of our proposed datasets (part 3).}
    \label{table:medium3}
    \def\arraystretch{1.2}
    \begin{center}
        \bgroup
        \setlength{\tabcolsep}{0.2em}
        \scriptsize
        \begin{tabular}{lcllcccc} 
            \toprule
            \multicolumn{1}{c}{\multirow{2}{*}{\bf Eq. ID}} & \multicolumn{1}{c}{\multirow{2}{*}{\bf Formula}} & \multicolumn{2}{c}{\multirow{2}{*}{\bf Symbols}} & \multicolumn{2}{c}{\bf Properties} & \multicolumn{2}{c}{\bf Distributions}
            \\ %
            & & & & Original & Ours & Original & Ours \\\midrule
            \multirow{5}{*}{I.43.43} & \multirow{5}{*}{$\kappa = \frac{1}{\gamma - 1} \frac{k v}{\sigma_c}$} & $\kappa$ & Thermal conductivity & V, F, P & V, F, P & N/A & N/A\\
             &  & $\gamma$ & Heat capacity ratio & V, F, P & V, F, P & $\U(2, 5)$ & $\U(1, 2)$\\
             &  & $k$ & Boltzmann constant & V, F, P & C, F, P & $\U(1, 5)$ & $1.381 \times 10^{-23}$\\
             &  & $v$ & Velocity & V, F, P & V, F, P & $\U(1, 5)$ & $\Ulog(10^{2}, 10^{4})$\\
             &  & $\sigma_c$ & Molecular collision cross section & V, F, P & V, F, P & $\U(1, 5)$ & $\Ulog(10^{-21}, 10^{-19})$\\\hline
            \multirow{4}{*}{I.48.2} & \multirow{4}{*}{$E = \frac{m c^2}{\sqrt{1-v^2/c^2}}$} & $E$ & Energy & V, F, P & V, F, P & N/A & N/A\\
\bottomrule
        \end{tabular}
        \egroup
    \end{center}
\end{table}

\begin{table}[H]
    \caption{\uline{Medium set} of our proposed datasets (part 4).}
    \label{table:medium4}
    \def\arraystretch{1.2}
    \begin{center}
        \bgroup
        \setlength{\tabcolsep}{0.1em}
        \scriptsize
        \begin{tabular}{lcllcccc} 
            \toprule
            \multicolumn{1}{c}{\multirow{2}{*}{\bf Eq. ID}} & \multicolumn{1}{c}{\multirow{2}{*}{\bf Formula}} & \multicolumn{2}{c}{\multirow{2}{*}{\bf Symbols}} & \multicolumn{2}{c}{\bf Properties} & \multicolumn{2}{c}{\bf Distributions}
            \\ %
            & & & & Original & Ours & Original & Ours \\\midrule
            \multirow{5}{*}{III.4.32} & \multirow{5}{*}{$n = \frac{1}{\exp(h \omega/2 \pi k T) - 1}$} & $n$ & Average number of photons & V, F, P & V, F, P & N/A & N/A\\
             &  & $h$ & Planck constant & V, F, P & C, F, P & $\U(1, 5)$ & $6.626 \times 10^{-34}$\\
             &  & $\omega$ & Frequency & V, F, P & V, F, P & $\U(1, 5)$ & $\Ulog(10^{8}, 10^{10})$\\
             &  & $k$ & Boltzmann constant & V, F, P & C, F, P & $\U(1, 5)$ & $1.381 \times 10^{-23}$\\
             &  & $T$ & Temperature & V, F, P & V, F, P & $\U(1, 5)$ & $\Ulog(10^{1}, 10^{3})$\\\hline
            \multirow{4}{*}{III.8.54} & \multirow{4}{*}{$|C|^2 = \sin^2\left(\frac{2 \pi A t}{h}\right)$} & $|C|^2$ & Probability & V, F & V, F, NN & N/A & N/A\\
\bottomrule
        \end{tabular}
        \egroup
    \end{center}
\end{table}

\begin{table}[H]
    \caption{\uline{Hard set} of our proposed datasets (part 1).}
    \label{table:hard1}
    \def\arraystretch{1.3}
    \begin{center}
        \bgroup
        \setlength{\tabcolsep}{0.1em}
        \tiny
        \begin{tabular}{lclp{10em}cccc} 
            \toprule
            \multicolumn{1}{c}{\multirow{2}{*}{\bf Eq. ID}} & \multicolumn{1}{c}{\multirow{2}{*}{\bf Formula}} & \multicolumn{2}{c}{\multirow{2}{*}{\bf Symbols}} & \multicolumn{2}{c}{\bf Properties} & \multicolumn{2}{c}{\bf Distributions}
            \\ %
            & & & & Original & Ours & Original & Ours \\\midrule
            \multirow{3}{*}{I.6.20} & \multirow{3}{*}{$f = \exp\left(-\frac{\theta^2}{2\sigma^2}\right)/\sqrt{2\pi\sigma^2}$} & $f$ & Probability density function & V, F, P & V, F, P & N/A & N/A\\
             &  & $\theta$ & Position & V, F, P & V, F & $\U(1, 3)$ & $\Ulog(10^{-1}, 10^{1})$\\
             &  & $\sigma$ & Standard deviation & V, F, P & V, F, P & $\U(1, 3)$ & $\Ulog(10^{-1}, 10^{1})$\\\hline
            \multirow{2}{*}{I.6.20a} & \multirow{2}{*}{$f = \exp\left(-\frac{\theta^2}{2}\right)/\sqrt{2\pi}$} & $f$ & Probability density function & V, F, P & V, F, P & N/A & N/A\\
             &  & $\theta$ & Position & V, F, P & V, F & $\U(1, 3)$ & $\Ulog(10^{-1}, 10^{1})$\\\hline
            \multirow{4}{*}{I.6.20b} & \multirow{4}{*}{$f = \exp\left(-\frac{\left(\theta-\theta_1\right)^2}{2\sigma^2}\right)/\sqrt{2\pi\sigma}$} & $f$ & Probability density function & V, F, P & V, F, P & N/A & N/A\\
\bottomrule
        \end{tabular}
        \egroup
    \end{center}
\end{table}

\begin{table}[H]
    \caption{\uline{Hard set} of our proposed datasets (part 2).}
    \label{table:hard2}
    \def\arraystretch{1.3}
    \begin{center}
        \bgroup
        \setlength{\tabcolsep}{0.2em}
        \tiny
        \begin{tabular}{lclp{12em}cccc} 
            \toprule
            \multicolumn{1}{c}{\multirow{2}{*}{\bf Eq. ID}} & \multicolumn{1}{c}{\multirow{2}{*}{\bf Formula}} & \multicolumn{2}{c}{\multirow{2}{*}{\bf Symbols}} & \multicolumn{2}{c}{\bf Properties} & \multicolumn{2}{c}{\bf Distributions}
            \\ %
            & & & & Original & Ours & Original & Ours \\\midrule
            \multirow{4}{*}{I.37.4} & \multirow{4}{*}{$\begin{aligned}I_{12} &= I_1+I_2\\&+2 \sqrt{I_1 I_2} \cos\delta\end{aligned}$} & $I_{12}$ & Amplitude of wave & V, F, NN & V, F, NN & N/A & N/A\\
             &  & $I_1$ & Amplitude of wave & V, F, P & V, F, P & $\U(1, 5)$ & $\Ulog(10^{-3}, 10^{-1})$\\
             &  & $I_2$ & Amplitude of wave & V, F, P & V, F, P & $\U(1, 5)$ & $\Ulog(10^{-3}, 10^{-1})$\\
             &  & $\delta$ & Phase difference & V, F, P & V, F, NN & $\U(1, 5)$ & $\U(0, 2 \pi)$\\\hline
            \multirow{5}{*}{I.39.22} & \multirow{5}{*}{$P = \frac{n k T}{V}$} & $P$ & Pressure & V, F, P & V, F, P & N/A & N/A\\
             &  & $n$ & Number of molecules & V, F, P & V, I$\star$, P & $\U(1, 5)$ & $\Ulog(10^{23}, 10^{25})$\\
\bottomrule
        \end{tabular}
        \egroup
    \end{center}
\end{table}

\begin{table}[H]
    \caption{\uline{Hard set} of our proposed datasets (part 3).}
    \label{table:hard3}
    \def\arraystretch{1.3}
    \begin{center}
        \bgroup
        \setlength{\tabcolsep}{0.2em}
        \tiny
        \begin{tabular}{lclp{10em}cccc} 
            \toprule
            \multicolumn{1}{c}{\multirow{2}{*}{\bf Eq. ID}} & \multicolumn{1}{c}{\multirow{2}{*}{\bf Formula}} & \multicolumn{2}{c}{\multirow{2}{*}{\bf Symbols}} & \multicolumn{2}{c}{\bf Properties} & \multicolumn{2}{c}{\bf Distributions}
            \\ %
            & & & & Original & Ours & Original & Ours \\\midrule
            \multirow{5}{*}{II.11.27} & \multirow{5}{*}{$P = \frac{N \alpha}{1-(N \alpha/3)} \epsilon E$} & $P$ & Polarizability & V, F, P & V, F & N/A & N/A\\
             &  & $N$ & Number of atom & V, F, NN & V, I$\star$, P & $\U(0, 1)$ & $\Ulog(10^{23}, 10^{25})$\\
             &  & $\alpha$ & Molecular polarizability & V, F, NN & V, F, P & $\U(0, 1)$ & $\Ulog(10^{-33}, 10^{-31})$\\
             &  & $\epsilon$ & Vacuum permittivity & V, F, P & C, F, P & $\U(1, 2)$ & $8.854 \times 10^{-12}$\\
             &  & $E$ & Magnitude of electric field & V, F, P & V, F, P & $\U(1, 2)$ & $\Ulog(10^{1}, 10^{3})$\\\hline
            \multirow{3}{*}{II.11.28} & \multirow{3}{*}{$\kappa = 1 + \frac{N \alpha}{1-(N \alpha/3)}$} & $\kappa$ & Electric dipole moment per unit volume & V, F, P & V, F & N/A & N/A\\
\bottomrule
        \end{tabular}
        \egroup
    \end{center}
\end{table}

\begin{table}[H]
    \caption{\uline{Hard set} of our proposed datasets (part 4).}
    \label{table:hard4}
    \def\arraystretch{1.3}
    \begin{center}
        \bgroup
        \setlength{\tabcolsep}{0.05em}
        \tiny
        \begin{tabular}{lcllcccc} 
            \toprule
            \multicolumn{1}{c}{\multirow{2}{*}{\bf Eq. ID}} & \multicolumn{1}{c}{\multirow{2}{*}{\bf Formula}} & \multicolumn{2}{c}{\multirow{2}{*}{\bf Symbols}} & \multicolumn{2}{c}{\bf Properties} & \multicolumn{2}{c}{\bf Distributions}
            \\ %
            & & & & Original & Ours & Original & Ours \\\midrule
            \multirow{5}{*}{III.10.19} & \multirow{5}{*}{$E = \mu \sqrt{B_x^2+B_y^2+B_z^2}$} & $E$ & Energy & V, F, P & V, F & N/A & N/A\\
             &  & $\mu$ & Magnetic moment & V, F, P & V, F & $\U(1, 5)$ & $\Ulog(10^{-25}, 10^{-23})$\\
             &  & $B_x$ & Element of magnetic field & V, F, P & V, F & $\U(1, 5)$ & $\Ulog(10^{-3}, 10^{-1})$\\
             &  & $B_y$ & Element of magnetic field & V, F, P & V, F & $\U(1, 5)$ & $\Ulog(10^{-3}, 10^{-1})$\\
             &  & $B_z$ & Element of magnetic field & V, F, P & V, F & $\U(1, 5)$ & $\Ulog(10^{-3}, 10^{-1})$\\\hline
            \multirow{5}{*}{III.21.20} & \multirow{5}{*}{$J = -\rho \frac{q}{m} A$} & $J$ & Electric current & V, F, N & V, F & N/A & N/A\\
\bottomrule
        \end{tabular}
        \egroup
    \end{center}
\end{table}

\begin{table}[H]
    \caption{\uline{Hard set} of our proposed datasets (part 5).}
    \label{table:hard5}
    \def\arraystretch{1.3}
    \begin{center}
        \bgroup
        \setlength{\tabcolsep}{0.05em}
        \tiny
        \begin{tabular}{lcllcccc} 
            \toprule
            \multicolumn{1}{c}{\multirow{2}{*}{\bf Eq. ID}} & \multicolumn{1}{c}{\multirow{2}{*}{\bf Formula}} & \multicolumn{2}{c}{\multirow{2}{*}{\bf Symbols}} & \multicolumn{2}{c}{\bf Properties} & \multicolumn{2}{c}{\bf Distributions}
            \\ %
            & & & & Original & Ours & Original & Ours \\\midrule
            \multirow{4}{*}{B10} & \multirow{4}{*}{$\cos\theta_1 = \frac{\cos\theta_2-v/c}{(1-v/c) \cos\theta_2}$} & $\cos\theta_1$ & Value & V, F & V, F & N/A & N/A\\
             &  & $\theta_2$ & Angle & V, F, P & V, F, NN & $\U(1, 3)$ & $\U(0, \pi)$\\
             &  & $v$ & Velocity & V, F, P & V, F & $\U(1, 3)$ & $\Ulog(10^{6}, 10^{8})$\\
             &  & $c$ & Speed of light & V, F, P & C, F, P & $\U(4, 6)$ & $2.998 \times 10^{8}$\\\hline
            \multirow{5}{*}{B11} & \multirow{5}{*}{$I = I_0 \left(\frac{\sin(\alpha/2)}{\alpha/2} \frac{\sin(N \delta/2)}{\sin(\delta/2)}\right)^2$} & $I$ & Wave intensity & V, F, P & V, F, P & N/A & N/A\\
             &  & $I_0$ & Amplitude of wave & V, F, P & V, F, P & $\U(1, 3)$ & $\Ulog(10^{-3}, 10^{-1})$\\
\bottomrule
        \end{tabular}
        \egroup
    \end{center}
\end{table}

\section{License of External Code}
\label{app_sec:external_code_license}
We briefly summarize the licenses of external code we used in this study.
BSD 3-Clause is used for both gplearn~\citep{koza2005genetic} (\url{https://gplearn.readthedocs.io/en/stable/index.html}), DSR~\citep{petersen2020deep} and uDSR~\citep{landajuela2022unified}(\url{https://github.com/brendenpetersen/deep-symbolic-optimization}).
Both AFP~\citep{schmidt2011age} and AFP-FE~\citep{schmidt2009distilling} (\url{https://github.com/cavalab/ellyn}) use GPL ver. 2 or later.
AIF~\citep{udrescu2020ai2} (\url{https://github.com/SJ001/AI-Feynman}) use MIT License, and both E2E~\citep{kamienny2022end} (\url{https://github.com/facebookresearch/symbolicregression}) and PySR~\citep{cranmer2023interpretable}(\url{https://github.com/MilesCranmer/PySR}) use Apache License 2.0.

\section{Hyperparameters for Symbolic Regression Baselines}
\label{app_sec:baseline_hyperparams}
Tables~\ref{table:hyperparams1} and~\ref{table:hyperparams2} show the hyperparameter space for symbolic regression baselines considered in this study.
The hyperparameters of gplearn~\citep{koza2005genetic}~\footnote{\url{https://gplearn.readthedocs.io/en/stable/reference.html\#symbolic-regressor}}, AFP~\citep{schmidt2011age}, and AFP-FE~\citep{schmidt2009distilling}~\footnote{\url{https://github.com/cavalab/ellyn}} are optimized by Optuna~\citep{akiba2019optuna}, a hyperparameter optimization framework.
For E2E~\citep{kamienny2022end}, we reuse the checkpoint of the pretrained model the authors provided.\footnote{\url{https://dl.fbaipublicfiles.com/symbolicregression/model1.pt}\label{fn:e2e_ckpt}}
We choose hyperparameters of other methods based on suggestions in their code and/or papers.

\begin{table}[H]
    \vspace{-1em}
    \caption{Hyperparameter sets for symbolic regression baselines (part 1).}
    \label{table:hyperparams1}
    \def\arraystretch{1.2}
    \begin{center}
        \bgroup
        \small
        \setlength{\tabcolsep}{0.2em}
        \begin{tabular}{c|l} 
            \toprule
            \multicolumn{1}{c|}{\bf Method} & \multicolumn{1}{c}{\bf Hyperparameter sets} \\
            \midrule
            gplearn & 100 trials with random combinations of the following hyperparameter spaces: \\
            & \emph{population\_size}: $\U(10^2, 10^3)$, \emph{generations}: $\U(10, 10^2)$, \\
            & \emph{stopping\_criteria}: $\U(10^{-10}, 10^{-2})$, \emph{warm\_start}: \{True, False\}, \\
            & \emph{const\_range}: \{None, $(-1.0, 1.0), (-10, 10), (-10^2, 10^2), (-10^3, 10^3), (-10^4, 10^4)$\}, \\
            & \emph{max\_samples}: $\U(0.9, 1.0)$, \emph{parsimony\_coefficient}: $\U(10^{-3}, 10^{-2})$ \\ \\
            AFP & 100 trials with random combinations of the following hyperparameter spaces: \\ 
            & \emph{popsize}: $\U(100, 1000)$, \emph{g}: $\U(250, 2500)$, \emph{stop\_threshold}: $\U(10^{-10}, 10^{-2})$, \\
            & \emph{op\_list}: \{['n', 'v', '+', '-', '*', '/', 'exp', 'log', '2', '3', 'sqrt'], \\
            & ['n', 'v', '+', '-', '*', '/', 'exp', 'log', '2', '3', 'sqrt', 'sin', 'cos']\} \\ \\
            AFP-FE & 100 trials with random combinations of the following hyperparameter spaces: \\ 
            & \emph{popsize}: $\U(100, 1000)$, \emph{g}: $\U(250, 2500)$, \emph{stop\_threshold}: $\U(10^{-10}, 10^{-2})$, \\
            & \emph{op\_list}: \{['n', 'v', '+', '-', '*', '/', 'exp', 'log', '2', '3', 'sqrt'], \\
            & ['n', 'v', '+', '-', '*', '/', 'exp', 'log', '2', '3', 'sqrt', 'sin', 'cos']\} \\ \\
            AIF & \{\emph{bftt}: 60, \emph{epoch}: 300, \emph{op}: '7ops.txt', \emph{poly\_deg}: 3\}, \\
            & \{\emph{bftt}: 60, \emph{epoch}: 300, \emph{op}: '10ops.txt', \emph{poly\_deg}: 3\}, \\ 
            & \{\emph{bftt}: 60, \emph{epoch}: 300, \emph{op}: '14ops.txt', \emph{poly\_deg}: 3\}, \\
            & \{\emph{bftt}: 60, \emph{epoch}: 300, \emph{op}: '19ops.txt', \emph{poly\_deg}: 3\}, \\
            & \{\emph{bftt}: 120, \emph{epoch}: 300, \emph{op}: '14ops.txt', \emph{poly\_deg}: 4\}, \\
            & \{\emph{bftt}: 120, \emph{epoch}: 300, \emph{op}: '19ops.txt', \emph{poly\_deg}: 4\}, \\
            & \{\emph{bftt}: 60, \emph{epoch}: 500, \emph{op}: '7ops.txt', \emph{poly\_deg}: 3\}, \\
            & \{\emph{bftt}: 60, \emph{epoch}: 500, \emph{op}: '10ops.txt', \emph{poly\_deg}: 3\}, \\ 
            & \{\emph{bftt}: 60, \emph{epoch}: 500, \emph{op}: '14ops.txt', \emph{poly\_deg}: 3\}, \\
            & \{\emph{bftt}: 60, \emph{epoch}: 500, \emph{op}: '19ops.txt', \emph{poly\_deg}: 3\} \\ \\
            DSR 
            & \{\emph{seed}: 1, \emph{function\_set}: ['add', 'sub', 'mul', 'div', 'sin', 'cos', 'exp', 'log'\}, \\ 
            & \{\emph{seed}: 2, \emph{function\_set}: ['add', 'sub', 'mul', 'div', 'sin', 'cos', 'exp', 'log'\}, \\ 
            & \{\emph{seed}: 3, \emph{function\_set}: ['add', 'sub', 'mul', 'div', 'sin', 'cos', 'exp', 'log'\}, \\ 
            & \{\emph{seed}: 4, \emph{function\_set}: ['add', 'sub', 'mul', 'div', 'sin', 'cos', 'exp', 'log'\}, \\ 
            & \{\emph{seed}: 5, \emph{function\_set}: ['add', 'sub', 'mul', 'div', 'sin', 'cos', 'exp', 'log'\}, \\ 
            & \{\emph{seed}: 1, \emph{function\_set}: ['add', 'sub', 'mul', 'div', 'sin', 'cos', 'exp', 'log', 'const']\}, \\
            & \{\emph{seed}: 2, \emph{function\_set}: ['add', 'sub', 'mul', 'div', 'sin', 'cos', 'exp', 'log', 'const']\}, \\ 
            & \{\emph{seed}: 3, \emph{function\_set}: ['add', 'sub', 'mul', 'div', 'sin', 'cos', 'exp', 'log', 'const']\}, \\ 
            & \{\emph{seed}: 4, \emph{function\_set}: ['add', 'sub', 'mul', 'div', 'sin', 'cos', 'exp', 'log', 'const']\}, \\ 
            & \{\emph{seed}: 5, \emph{function\_set}: ['add', 'sub', 'mul', 'div', 'sin', 'cos', 'exp', 'log', 'const']\} \\ \\
            E2E & We reused the checkpoint of the pretrained model the authors provided.$\textsuperscript{\ref{fn:e2e_ckpt}}$\\
            \bottomrule
        \end{tabular}
        \egroup
    \end{center}
\end{table}

\begin{table}[H]
    \vspace{-1em}
    \caption{Hyperparameter sets for symbolic regression baselines (part 2).}
    \label{table:hyperparams2}
    \def\arraystretch{1.2}
    \begin{center}
        \bgroup
        \small
        \setlength{\tabcolsep}{0.1em}
        \begin{tabular}{c|l} 
            \toprule
            \multicolumn{1}{c|}{\bf Method} & \multicolumn{1}{c}{\bf Hyperparameter sets} \\
            \midrule
            uDSR 
            & \{\emph{seed}: 1, \emph{function\_set}: ['add', 'sub', 'mul', 'div', 'sin', 'cos', 'exp', 'log', 'poly'], \\
            & \emph{batch\_size}: 1000, \emph{learning\_rate}: 0.0005, \emph{entropy\_weight}: 0.03\}, \\ 
            & \{\emph{seed}: 2, \emph{function\_set}: ['add', 'sub', 'mul', 'div', 'sin', 'cos', 'exp', 'log', 'poly'], \\
            & \emph{batch\_size}: 1000, \emph{learning\_rate}: 0.0005, \emph{entropy\_weight}: 0.03\}, \\ 
            & \{\emph{seed}: 3, \emph{function\_set}: ['add', 'sub', 'mul', 'div', 'sin', 'cos', 'exp', 'log', 'poly'], \\
            & \emph{batch\_size}: 1000, \emph{learning\_rate}: 0.0005, \emph{entropy\_weight}: 0.03\}, \\ 
            & \{\emph{seed}: 4, \emph{function\_set}: ['add', 'sub', 'mul', 'div', 'sin', 'cos', 'exp', 'log', 'poly'], \\
            & \emph{batch\_size}: 1000, \emph{learning\_rate}: 0.0005, \emph{entropy\_weight}: 0.03\}, \\ 
            & \{\emph{seed}: 5, \emph{function\_set}: ['add', 'sub', 'mul', 'div', 'sin', 'cos', 'exp', 'log', 'poly'], \\
            & \emph{batch\_size}: 1000, \emph{learning\_rate}: 0.0005, \emph{entropy\_weight}: 0.03\}, \\ 
            & \{\emph{seed}: 6, \emph{function\_set}: ['add', 'sub', 'mul', 'div', 'sin', 'cos', 'exp', 'log', 'poly'], \\
            & \emph{batch\_size}: 500, \emph{learning\_rate}: 0.0025, \emph{entropy\_weight}: 0.3\}, \\ 
            & \{\emph{seed}: 7, \emph{function\_set}: ['add', 'sub', 'mul', 'div', 'sin', 'cos', 'exp', 'log', 'poly'], \\
            & \emph{batch\_size}: 500, \emph{learning\_rate}: 0.0025, \emph{entropy\_weight}: 0.3\},  \\ 
            & \{\emph{seed}: 8, \emph{function\_set}: ['add', 'sub', 'mul', 'div', 'sin', 'cos', 'exp', 'log', 'poly'], \\
            & \emph{batch\_size}: 500, \emph{learning\_rate}: 0.0025, \emph{entropy\_weight}: 0.3\}, \\ 
            & \{\emph{seed}: 9, \emph{function\_set}: ['add', 'sub', 'mul', 'div', 'sin', 'cos', 'exp', 'log', 'poly'], \\
            & \emph{batch\_size}: 500, \emph{learning\_rate}: 0.0025, \emph{entropy\_weight}: 0.3\},  \\ 
            & \{\emph{seed}: 10, \emph{function\_set}: ['add', 'sub', 'mul', 'div', 'sin', 'cos', 'exp', 'log', 'poly'], \\
            & \emph{batch\_size}: 500, \emph{learning\_rate}: 0.0025, \emph{entropy\_weight}: 0.3\} \\ \\
            PySR & \emph{procs}: 5, \emph{populations}: 10, \emph{population\_size}: 40, \emph{ncyclesperiteration}: 500, \\
            & \emph{niterations}: 50000, \emph{timeout\_in\_seconds}: 82800, \emph{maxsize}: 50, \\
            & \emph{binary\_operators}: ['*', '+', '-', '/'], \emph{unary\_operators}: ['sin', 'cos', 'exp', 'log'], \\
            & \emph{nested\_constraints}: \{sin: \{sin: 0, cos: 0\}, cos: \{sin: 0, cos: 0\}, \\
            & exp: \{exp: 0\}, log: \{log: 0\}\}, \emph{progress}: False, \emph{weight\_randomize}: 0.1, \\
            & \emph{precision}: 32, \emph{warm\_start}: False, \emph{turbo}: True, \emph{update}: False
            \\
            \bottomrule
        \end{tabular}
        \egroup
    \end{center}
\end{table}

\section{Qualitative Analysis}
\label{app_sec:qualitative_analysis}
This section discusses qualitative analysis for the experimental results in Section~\ref{subsec:results}, focused on the effect of introduced dummy variables on the behaviors of SR baseline methods since the \ourproblem-Feynman datasets with dummy variables seems extremely challenging from the solution rate and NED in Tables~\ref{table:baseline_results} and~\ref{table:baseline_results_w_dummy_vars}.
Taking two \ourproblem problems (I.12.1 and II.27.16) as examples, Table~\ref{table:examples} highlights how the randomly introduced dummy variables made changes in both the true models (equations) and the SR baseline methods' predicted equations.
For the \ourproblem problem I.12.1 in Table~\ref{table:easy1}, all the SR baselines except E2E made the perfect predictions, which completely match the true model.
When introducing a random dummy variable ($x_2$ for this problem) to I.12.1, however, gplearn failed to complete the training process, and AIF, DSR, E2E, and uDSR produced little bit overcomplex symbolic expressions, including the dummy variable ($x_2$). 
A similar trend can be confirmed for another \ourproblem problem II.27.16 in Table~\ref{table:easy3}.
While AFP, AFP-FE, AIF, uDSR, and PySR produced the correct symbolic expressions in terms of NED (\emph{i.e.}, NED = 0), randomly introduced dummy variables $x_0$ and $x_1$ worsened the predictions of AFP-FE, AIF, and uDSR.
Note that even though the original \ourproblem problem II.27.16 contains only one input variable $x_0$, two dummy variables randomly introduced as the first and second columns of the tabular dataset reindexed the original input variable $x_0$ as $x_2$ in this specific dataset due to the dummy variables.

\begin{table}[t]
    \caption{Examples: \ourproblem problems I.12.1 (top) and II.27.16 (bottom) from \ourproblem-Feynman (\emph{Easy set}) to highlight how introduced dummy variables affected behaviors of the SR baselines. Coefficients are rounded for better presentation. N/A: No prediction obtained as the training process did not complete.}
    \label{table:examples}
    \def\arraystretch{1.2}
    \begin{center}
        \begin{tabularx}{\textwidth}{c|X|X}
            \toprule
            Dummy var(s). ? & \multicolumn{1}{c|}{\bf No} & \multicolumn{1}{c}{$x_2$} \\
            \midrule
            True model & $x_0 \cdot x_1$ & $x_0 \cdot x_1$ \\
            \midrule
            gplearn & $x_0 \cdot x_1$ & \nacell \\
            AFP & $x_0 \cdot x_1$ & $x_0 \cdot x_1$ \\
            AFP-FE & $x_0 \cdot x_1$ & $x_0 \cdot x_1$ \\
            AIF & $x_0 \cdot x_1$ & $0.999 \cdot x_0 \cdot x_1 + 0.159 \cdot x_2$ \\
            DSR & $x_0 \cdot x_1$ & $x_0 \cdot x_1 \cdot \exp\left(-x_2^2/\left(x_1 + x_2\right)\right) \cdot \cos\left(x_0 \cdot x_2\right)$ \\
            E2E & $1.02 \cdot \left(x_0 - 1.16\text{e-3}\right) \cdot \left(x_1 - 6.88\text{e-3}\right)$ & $1.32\text{e+17} \cdot \left(x_1 - 4.89\text{e-3}\right) \cdot \left(7.38\text{e-18} \cdot x_0 + x_2 + 1.11\text{e-20}\right)$ \\
            uDSR & $x_0 \cdot x_1$ & $x_1 \cdot (x_0 - \sin(x_2 \cdot \exp(x_2)/(x_0 - x_2)))$ \\
            PySR & $x_0 \cdot x_1$ & $x_0 \cdot x_1$ \\
            \midrule\midrule
            Dummy var(s). ? & \multicolumn{1}{c|}{\bf No} & \multicolumn{1}{c}{$x_0$, $x_1$} \\
            \midrule
            True model & $2.65\text{e-3} \cdot x_0^2$ & $2.65\text{e-3} \cdot x_2^2$ \\
            \midrule
            gplearn & \nacellmid & \nacell \\
            AFP & $2.68\text{e-3} \cdot x_0^2$ & $2.66\text{e-3} \cdot x_2^2$ \\
            AFP-FE & $2.65\text{e-3} \cdot x_0^2$ & $2.10\text{e-3} \cdot x_2^2 - 0.0161$ \\
            AIF & $2.65\text{e-3} \cdot x_0^2$ & \nacell \\
            DSR & \nacellmid & $4.91\text{e-3} \cdot x_2^2 \cdot \cos\left(\exp\left(x_0\right)\right)$ \\
            E2E & $(0.191 \cdot x_0 - 0.0375) \cdot (\tan(0.0137 \cdot x_0 + 3.21\text{e-3}) - 7.05\text{e-5})$ & $(2.60\text{e-3} \cdot x_2 + 4.75\text{e-5}) \cdot (x_2 - 0.0104 \cdot \sin(1.84\text{e+21} \cdot x_0 - 2.40\text{e-11} \cdot x_1 - 9.12\text{e+26} \cdot x_3 + 1.36) + 0.0398)$ \\
            uDSR & $2.65\text{e-3} \cdot x_0^2$ & $2.65\text{e-3} \cdot x_1 \cdot x_2^2 / (x_1 - x_2)$ \\
            PySR & $2.65\text{e-3} \cdot x_0^2$ & $2.65\text{e-3} \cdot x_0^2$  \\
            \bottomrule
        \end{tabularx}
    \end{center}
\end{table}

\section{Injecting Noise to Target Variables}
\label{app_sec:noise_injection}
Following SRBench~\citep{la2021contemporary}, we introduce Gaussian noise with a parameter of noise level $\gamma$ to the target variables in our \ourproblem datasets.
We inject the noise to each of the datasets separately (Eq. (\ref{eq:target_noise})):
\begin{equation}
    y_{j}^\text{noise} = f_\text{true}\left( X_j \right) + \epsilon,~~~~~ \epsilon \sim \mathcal{N}\left(0, \gamma \sqrt{\frac{1}{N} \sum_{k=1}^{N} f_\text{true}\left( X_k \right)}\right),
    \label{eq:target_noise}
\end{equation}
\noindent where $1 \leq j \leq N$ and $N$ indicates the number of samples in the dataset.

Table~\ref{table:baseline_w_noise_edit_distance} shows normalized edit distances of our baselines for noise-injected \ourproblem (Easy), reusing the set of noise levels in SRBench~\citep{la2021contemporary} \emph{i.e.}, $\gamma \in \{0, 10^{-3}, 10^{-2}, 10^{-1}\}$.
Overall, the more the injected noise is, the more difficult it would be for the baseline models to (re-)discover the physical law in the data.

\begin{table}[t]
    \begin{center}
        \caption{Normalized edit distances of baselines for \uline{noise-injected \ourproblem (Easy) datasets} with different noise levels.}
        \label{table:baseline_w_noise_edit_distance}
        \vspace{0.5em}
        \begin{tabular}{c|rrrrr} 
            \toprule
            \multicolumn{1}{c|}{\bf Noise Level ($\gamma$) \textbackslash~Method} & \multicolumn{1}{c}{\bf gplearn} & \multicolumn{1}{c}{\bf AFP} & \multicolumn{1}{c}{\bf AFP-FE} & \multicolumn{1}{c}{\bf AIF} & \multicolumn{1}{c}{\bf DSR} \\
            \midrule
            0 & 0.876 & 0.703 & 0.712 & 0.646 & \highlight 0.551  \\ 
            $10^{-3}$ & 0.928 & 0.799 & 0.814 & \highlight 0.797 & 0.820 \\ 
            $10^{-2}$ & 0.940 & 0.824 & 0.880 & 0.870 & \highlight 0.793 \\ 
            $10^{-1}$ & 0.948 & \highlight 0.823 & 0.960 & 0.882 & 0.841 \\ 
            \bottomrule
        \end{tabular}
    \end{center}
    \begin{center}
        \caption{Solution rates of the common baselines between \orgdataset and \ourproblem-Feynman datasets.}
        \label{table:solution_rate_fsrd_vs_srsd}
        \vspace{0.5em}
        \bgroup
        \setlength{\tabcolsep}{0.4em}
        \begin{tabular}{c|rrrrr} 
            \toprule
            \multicolumn{1}{c|}{\bf Dataset \textbackslash~Method} & \multicolumn{1}{c}{\bf gplearn} & \multicolumn{1}{c}{\bf AFP} & \multicolumn{1}{c}{\bf AFP-FE} & \multicolumn{1}{c}{\bf AIF} & \multicolumn{1}{c}{\bf DSR} \\
            \midrule
            \orgdataset~\citep{udrescu2020ai} & 15.7\% & 20.41\% & 26.08\% & \highlight 53.0\% & 19.1\% \\ 
            \ourproblem (Ours) & 1.67\% & 5.83\% & 6.67\% & 9.17\% & \highlight 15.8\% \\ 
            \bottomrule
        \end{tabular}
        \egroup
    \end{center}
\end{table}

\section{Solution Rate Comparison - \orgdataset vs. \ourproblem~-}
\label{app_sec:solution_rate_fsrd_vs_srsd}

Table~\ref{table:solution_rate_fsrd_vs_srsd} compares the solution rates of the five common baselines for the \orgdataset and our \ourproblem datasets.
We can confirm that the overall solution rates for our \ourproblem are significantly degraded compared to those for the \orgdataset datasets reported in SRBench~\citep{la2021contemporary} except for DSR.\footnote{Chi-squared tests showed p-values of $4.30 \times 10^{-5}$, $1.05 \times 10^{-4}$, $1.61 \times 10^{-6}$, $1.99 \times 10^{-21}$, and $0.479$ for gplearn, AFP, AFP-FE, AIF, and DSR respectively.}
The results indicate that our \ourproblem datasets are more challenging than the \orgdataset datasets in terms of solution rate.

\section{Limitations}
\label{app_sec:limitations}

\subsection{Implicit Functions}
Symbolic regression generally has a limitation in inferring implicit functions, as the model infers a trivial constant function if there are no restrictions on variables. For example, $f(x,y)=0$ is inferred as $0=0 \ \forall x,y$. This problem can be solved by applying the constraint that an inferred function should depend on at least two variables \emph{e.g.}, inferring $f(x, y)=0$ with $\frac{\partial f}{\partial x} \neq 0$ and $\frac{\partial f}{\partial y} \neq 0$, or by converting the function to an explicit form \emph{e.g.}, $y=g(x)$. We converted some functions in the datasets into explicit forms and avoided the inverse trigonometric functions as described in Section~\ref{subsec:dataset_issues}.

\subsection{Noise Injection}
When applying machine learning to real-world problems, it is often true that the observed values contain some noise.
While we follow~\citet{la2021contemporary} and show experimental results for our \ourproblem datasets with noise-injected target variables in Appendix~\ref{app_sec:noise_injection}, these aspects are not thoroughly discussed in this study, such discussions can be a separate paper built on this work and further engage studies of symbolic regression for scientific discovery.

\subsection{Dependency on \textsf{sympy}}
Similar to SRBench~\citep{la2021contemporary}, the implementation of our evaluation pipeline has a significant dependency on \textsf{sympy}.
Specifically, when computing edit distance between the predicted and true expressions and solution rate, our evaluation pipeline builds equation trees based on the tree structure of the expressions used in \textsf{sympy} after converting the expressions to floating-point approximations.
Our use of edit distance and solution rate is based on our observation and an assumption that \textsf{sympy} consistently maps a given equation to the unique equation tree, handling algebraic properties so that we can compute edit distance between the true and estimated equation trees consistently.
We also acknowledge that \textsf{sympy} may fail to process too complex expressions, and some symbolic regression methods may produce such solutions.
However, since the interpretability of the prediction is a key property of symbolic regression, such overcomplex expressions should not be desired by non-ML users and will result in NED = 1 for \ourproblem problems considered in this study.

\end{document}